\documentclass[conference]{IEEEtran}
\IEEEoverridecommandlockouts
\usepackage{cite}
\usepackage{amsmath,amssymb,amsfonts}
\usepackage[table,xcdraw]{xcolor}
\usepackage{algorithm}
\usepackage{algpseudocode}
\usepackage{graphicx}
\usepackage{textcomp}
\usepackage{bbm}
\usepackage{xcolor}
\usepackage{geometry}
\usepackage{enumerate}
\usepackage{url}
\graphicspath{{./Images/}}
\usepackage{mdframed}
\usepackage[shortlabels]{enumitem}
\usepackage{optidef}
\usepackage[normalem]{ulem}
\usepackage{graphicx} 
\usepackage{algorithm}
\usepackage{algpseudocode}
\usepackage{mathrsfs}
\usepackage{amssymb}
\usepackage{lipsum}
\usepackage{amsfonts} 
\usepackage{amsmath}
\usepackage{hyperref}
\usepackage{amsthm}
\usepackage{color}
\usepackage{etoolbox}
\usepackage{epstopdf}
\usepackage{mathtools}
\usepackage{cite}
\usepackage{hyperref}
\def\BibTeX{{\rm B\kern-.05em{\sc i\kern-.025em b}\kern-.08em
    T\kern-.1667em\lower.7ex\hbox{E}\kern-.125emX}}
\usepackage{subcaption}
\newtheorem{lem}{Lemma}

\begin{document}

\title{Constrained Best Arm Identification \\ in Grouped Bandits
}

\newtheorem{theorem}{Theorem}
\newtheorem{corollary}{Corollary}
\newtheorem{lemma}{Lemma}
\newtheorem{remark}{Remark}

\author{\IEEEauthorblockN{Sahil Dharod, Malyala Preethi Sravani, Sakshi Heda, and Sharayu Moharir}
\IEEEauthorblockA{\textit{Department of Electrical Engineering, IIT Bombay}\\
\textit{210070026@iitb.ac.in, preethimalyala@gmail.com, sakshiheda.iitb@gmail.com, sharayum@ee.iitb.ac.in}
}
}

\maketitle

\begin{abstract}
We study a grouped bandit setting where each arm comprises multiple independent sub-arms referred to as attributes. Each attribute of each arm has an independent stochastic reward. We impose the constraint that for an arm to be deemed feasible, the mean reward of all its attributes should exceed a specified threshold.
The goal is to find the arm with the highest mean reward averaged across attributes among the set of feasible arms in the fixed confidence setting. We first characterize a fundamental limit on the performance of any policy. Following this, we propose a near-optimal confidence interval-based policy to solve this problem and provide analytical guarantees for the policy. We compare the performance of the proposed policy with that of two suitably modified versions of action elimination via simulations.

\end{abstract}


\section{Introduction}

Many amenities are a collection of independent services and each customer of such a service may use one or more at a time. For instance, a typical auto garage may offer car wash services, AC servicing, tyre and wheel care services, car inspections, etc. Similarly, a typical salon offers hair services, skin services, nail services, make-up services, etc. To evaluate such services, it makes sense to have customers rate each service separately and maintain a rating for each service. A reasonable metric for evaluating such a service as a whole is the (weighted) average of the ratings of the different services. In addition, for an amenity to be deemed acceptable, it may be desirable that the ratings for each service exceed a threshold. 

Motivated by this, we consider a grouped bandit setting where each arm is a group of independent sub-arms. We refer to these sub-arms as attributes. All the attributes of all arms are modeled as an independent stochastic process with a corresponding mean reward. A group is said to be feasible if the mean reward of all its attributes exceeds a given threshold. In this work, we focus on the problem of identifying the feasible group with the highest mean reward averaged across all its attributes in the fixed confidence setting \cite{track-and-stop}. We consider the setting where the learning agent can choose to sample a specific attribute of a specific arm. 

\subsection{Our Contributions} 
We first characterize a lower bound on the performance of any online policy. 
We then propose a policy that determines which arm-attribute pair(s) to sample at each round. We provide analytical performance guarantees for this policy and show that it is near-optimal. Further, we empirically compare the performance of the proposed policy with suitably adapted versions of a widely studied policy -- action elimination. Our numerical results show that our algorithm outperforms these other algorithms.

\color{black}
\subsection{Related Work} The best arm identification problem for multi-armed bandits has been widely studied. The two settings of this problem that have received the most attention are the fixed confidence setting \cite{track-and-stop} and the fixed budget setting \cite{fixed-budget}. As mentioned above, in this work, we focus on the former. In the fixed confidence setting, we are given a confidence parameter $\delta \in [0,1]$ as an input parameter. The goal of the learner is to identify the best arm with a probability of at least $1-\delta$ using as few samples of the arms as possible. A detailed survey of various algorithms proposed for this setting can be found in \cite{cbb-algos}. 

The two key features of the problem we are interested in are: \emph{(i)} each arm is a group of independent sub-arms, and \emph{(ii)} identifying the best arm among those which satisfy a ``feasibility'' constraint. We now discuss existing literature that looks at the best arm identification problem with these two features.

A version of the grouped arm problem is the focus of \cite{max-min-grouped-bandits}, where the goal is to identify the group with the highest minimum mean reward. Another grouped arm problem is studied in \cite{grouped-bai}, where authors discuss an algorithm to identify the best arm in each group. In \cite{best-m-arms}, the authors lay emphasis on identifying the best $m$ arms that attain the highest rewards out of the group of arms. Another related problem is called the categorized multi-armed bandits \cite{categorized-bandits}. Here, arms are grouped into different categories with an existing order between these categories, and the knowledge of the group structure is known. The goal is to find the overall best arm.

Multiple works have added a feasibility constraint to the best arm identification problem, where the learning agent is expected to select an optimal arm that satisfies a feasibility rule. In \cite{Safety-constraints}, the authors consider the best arm identification problem with linear and then monotonic safety constraints. A method to solve a feasibility-constrained best-arm identification (FC-BAI) problem for a general feasibility rule is shown in \cite{top-feasible-arm-identification}. In \cite{subpopulations}, the authors solve the FC-BAI problem with a constraint on arm mean using the track and stop technique \cite{track-and-stop}. In \cite{va-lucb}, the feasibility constraint is in terms of the variance of an arm. Under this constraint, an arm is feasible if its variance is below a given threshold. We use ideas from \cite{va-lucb} in the design and analysis of our algorithm. 

\color{black}
\section{Problem Setting}
There are $N$ arms and $[N] = \{1,2,\cdots, N\}$ denotes the set of all arms. Each arm has $M$ attributes, denoted by $[M] = \{1,2,\cdots, M\}.$ The reward corresponding to each attribute of each arm is modelled as an independent stochastic process. 
We use $\nu_{ij}$ to denote the distribution of attribute $j \in [M]$ of arm $i \in [N].$ 
The reward of attribute $j$ of arm $i$ in round $t$ is a stochastic random variable denoted by $X_{ij}(t) \sim \nu_{ij}.$ We define $\mu_{ij} = \mathbb{E}[X_{ij}(t)].$


Further, we are given a threshold, $\mu_{\text{TH}}$, which determines the feasibility of each arm. An arm $i$ is said to be feasible if and only if the mean reward of each of its attributes is at least $\mu_{\text{TH}}.$ 
We define the set of feasible arms, referred to as \emph{Feasible Set}, denoted by $\mathcal{F}$ as follows:
\begin{equation}
    \mathcal{F}\coloneqq\{i \in [N]: \min_j \mu_{ij} \geq \mu_{\text{TH}}\}.
\end{equation}

A problem instance is said to be feasible if $\mathcal{F} \neq \emptyset$ and is called infeasible otherwise. We define the feasibility flag $f$ as follows:
\begin{align*}
    f \coloneqq \begin{cases} 
        1 & \text{if the instance is feasible} \\
        0 & \text{otherwise.} 
    \end{cases}  
\end{align*}
For a feasible instance, the best arm $i^{\star}$ is defined as the arm with the highest mean reward, $\mu_i$ in $\mathcal{F}.$ An arm's mean reward, here denotes the average of the mean rewards of all of its attributes. i.e., 
\begin{align}
   i^{\star} \coloneqq & \arg\max_{i \in \mathcal{F}} \mu_i,  \
   \text{where, } \mu_i \coloneqq \left(\sum_{j=1}^M \mu_{ij}\right)/M. \nonumber
\end{align}
For example, consider the problem instance in Table \ref{tab:example}. Here, we have three arms, each with two attributes. Let $\mu_{\text{TH}} = 0.3.$ In this case, Arm 1 and Arm 3 are feasible, with Arm 1 being the best feasible arm. Arm 2 has the highest   mean reward but is infeasible since the mean reward for Attribute 1 is less than $\mu_{\text{TH}}.$



\begin{table}
    \centering    
    \caption{Illustrative Example}
    \label{tab:example}
    \begin{tabular}{|c|c|c|c|}
        \hline
        \textbf{Arm} & \textbf{Attribute 1} & \textbf{Attribute 2} & \textbf{Mean Reward} \\
        \hline
        1 & 0.6 & 0.4 & 0.5 \\
        \hline
        2 & 0.2 & 1 & 0.6 \\
        \hline
        3 & 0.4 & 0.4 & 0.4 \\
        \hline
    \end{tabular}
\end{table}


We assume that the best arm, if it exists, is unique. The rewards are considered to be bounded in $[0, 1]$.
The algorithmic challenge for the learner is to decide which arm-attribute pairs to play in each round. Table \ref{tab:notation_ps} summarizes the notation used in the section.




\begin{table}
    \centering
    \caption{Notation}
    \label{tab:notation_ps}
    \begin{tabular}{|c|c|}
        \hline
        \textbf{Notation} & \textbf{Description} \\
        \hline
        $X_{ij}(t)$ & Reward for arm $i,$ attribute $j$ in round $t$ \\
        \hline
        $\mu_{ij}$ & Mean reward of arm $i,$ attribute $j$ \\
        \hline
        $\mu_i$ & Mean reward of arm $i$ \\
        \hline
        $\mathcal{F}$ & Feasible Set \\
        \hline
        $\mu_{\text{TH}}$ & Threshold \\
        \hline
        $f$ & Feasibility flag \\
        \hline
        $i^{\star}$ & Index of the best feasible arm \\
        \hline
    \end{tabular}
    
\end{table}

\section{Lower Bound}
In this section, we provide a fundamental lower bound on the expected cost incurred by any online policy. 
\begin{theorem}[Lower bound]\label{th: lower bound}
    Given any instance $ (\nu,\mu_{\text{TH}}) $, we define a constant $C(\nu,\mu_{\text{TH}})$ and the lower bound on the expected sample complexity $\tau_\delta^\star$ in terms of the hardness index $(H_{\text{id}})$ is given by :

    \begin{align}
		&\tau_\delta^\star
		\ge  
		\ln \bigg( \frac{1}{2.4\delta} \bigg) C(\nu,\mu_{\text{TH}})\, H_\mathrm{id}.  
\end{align} 

\end{theorem}
A detailed proof is given in the appendix below.
\section{Our Algorithm: Confidence Set Sampling}\label{sec: algo}

We propose an LUCB-style algorithm \cite{va-lucb}, where we sample multiple arms in each round. Note that we have an extra degree of freedom as we can choose the attributes to be sampled.
We divide the arm-attribute pairs into subsets based on their potential feasibility and explore arms for which we are still determining the feasibility. We also explore the arms with assured feasibility to get tighter confidence bounds for their mean rewards. We stop when we ascertain that the mean reward of the current best feasible arm is greater than that of any other feasible arm.


The algorithm starts with a uniform exploration for each attribute of each arm. Rounds are indexed by $t,$ and the total number of pulls till round $t$ is denoted by $k(t).$ Let $\mathcal{J}_t$ denote the set of arm-attribute pairs pulled in round $t.$ We define $T_{ij}(t)$ as the number of samples of attribute $j$ of arm $i$ taken till round $t.$ Similarly, $T_i(t)$ is the number of times the least explored attribute of arm $i$ is sampled till time $t.$ Formally,
\begin{align*}
    T_{ij}(t) \coloneqq \sum \limits_{s=1}^{t-1} \mathbbm{1}_{(i,j) \in \mathcal{J}_s},\ T_i(t) \coloneqq \min \limits_{j\in [M]} T_{ij}(t).
\end{align*}
The empirical mean of the reward of attribute $j$ of arm $i$ is denoted by $\hat{\mu}_{ij}(t).$ The empirical average reward of arm $i$ is denoted by $\hat{\mu}_i(t).$ Formally,
\begin{equation}\label{eq: emp_mean}
\begin{split}
\hat{\mu}_{ij}(t) \coloneqq& \frac{1}{T_{ij}(t)}\sum_{s=1}^{t-1}X_{ij}(s)\mathbbm{1}_{(i, j) \in \mathcal{J}_s},\\
\hat{\mu}_i(t)\coloneqq& \left(\sum \limits_{j=1}^{M}\hat{\mu}_{ij}(t)\right)/{M}.
\end{split}
\end{equation}
Confidence radii for the arms and attributes is defined as 
\begin{equation*}
    \alpha(t, T(t))\coloneqq\sqrt{\frac{1}{2T(t)}\ln\left(\frac{4NMt^4}{\delta}\right)},
\end{equation*}
where $\delta$ is the confidence parameter and $T(t)$ corresponds to the arm or arm-attribute pair for which we are calculating the confidence radii.

We define the confidence intervals for each attribute with the lower confidence bound (LCB, denoted by $L_{ij}(t)$) and the upper confidence bound (UCB, denoted by $U_{ij}(t)$) as follows:
\begin{equation}\label{eq: bound_attr}
\begin{split}
L_{ij}(t)\coloneqq \hat{\mu}_{ij}(t) - \alpha(t, T_{ij}(t), k(t)),\\
U_{ij}(t)\coloneqq \hat{\mu}_{ij}(t) + \alpha(t, T_{ij}(t), k(t)).
\end{split}
\end{equation}

Similarly, we define the confidence interval for arm $i$ with the lower confidence bound (LCB, denoted by $L_{i}(t)$) and the upper confidence bound (UCB, denoted by $U_{i}(t)$) as follows:
\begin{equation}\label{eq: bound_arm}
\begin{split}
L_{i}(t)\coloneqq \hat{\mu}_{i}(t) - \alpha(t, T_i(t), k(t)),\\
U_{i}(t)\coloneqq \hat{\mu}_{i}(t) + \alpha(t, T_i(t), k(t)).
\end{split}
\end{equation}

Based on these confidence intervals, we define the following subsets for the arm-attribute pairs for round $t$:
\begin{enumerate}
    \item \emph{Perfectly Feasible Attribute Set:} the set of arm-attribute pairs whose lower confidence bound is above the threshold $\mu_{\text{TH}}.$ Formally,
\begin{equation*}
        \mathcal{F}_{Pt}^A \coloneqq \{ (i,j) \in [N] \times [M] : L_{ij}(t) \ge \mu_{\text{TH}}\}.
\end{equation*}
\item \emph{Almost Feasible Attribute Set:} the set of arm-attribute pairs whose lower confidence bound is less than the threshold $\mu_{\text{TH}}$ and the upper confidence bound is higher than the threshold $\mu_{\text{TH}}.$ In other words, the threshold lies within the confidence interval for these arm-attribute pairs. Formally,
\begin{equation*}
    \partial\mathcal{F}_t^A \coloneqq \{(i,j) \in [N] \times [M] : L_{ij}(t) < \mu_{\text{TH}} \le U_{ij}(t)\}.
\end{equation*}
\item \emph{Feasible Attribute Set:} the union of the \emph{Perfectly Feasible Attribute Set} and the \emph{Almost Feasible Attribute Set} and is denoted by $\mathcal{F}_t^A.$
\end{enumerate}


Based on the confidence intervals, we define the following subsets of the set of all arms:
\begin{enumerate}
    \item \emph{Perfectly Feasible Set:} the set of arms with all the attributes in the \emph{Perfectly Feasible Attribute Set}. Formally, 
    \begin{equation*}
        \mathcal{F}_{Pt} \coloneqq \{ i \in [N] : (i, j) \in \mathcal{F}_{Pt}^A, \ \forall j \in [M]\}.
    \end{equation*}
    \item \emph{Feasible Set:} the set of arms with all the attributes in the \emph{Feasible Attribute Set}, i.e., all the attributes have UCB higher than $\mu_{\text{TH}}$. Formally, 
    \begin{equation*}
        \mathcal{F}_t \coloneqq \{i \in [N]: (i, j) \in \mathcal{F}_t^A, \ \forall j \in [M]\}.
    \end{equation*}
    \item \emph{Almost Feasible Set:} the set of arms which are in the \emph{Feasible Set} but not in the \emph{Perfect Feasible Set} and is denoted by $\partial\mathcal{F}_t$.
    \item \emph{Potential Set:} the set of arms with UCB higher than the LCB of the empirically best arm. Formally, 
    \begin{equation}\label{eq: potential set}
        \mathcal{P}_t \coloneqq \begin{cases} 
            \{i \in [N] : L_{i^{\star}_t}(t) \le U_i(t)\} & \mathcal{F} \neq \phi \\
            [N] & \mathcal{F} = \phi,
        \end{cases}
    \end{equation}
    where $i^{\star}_t$ is the best arm from \emph{Perfect Feasible Set}
\end{enumerate}





These notations are summarized in Table \ref{tab:notation_algo}\footnote{Parameters that evolve over rounds are defined using the notation $*(t)$ or $*_t$.  }.
\begin{table}[!h]
    \centering
    \caption{Notation used in algorithm.}
    \label{tab:notation_algo}
    \begin{tabular}{|c|c|}
        \hline
        \textbf{Notation} & \textbf{Description} \\
        \hline
        $k(t)$ & Total number of pulls 
        \\\hline
        $T_{ij}(t)$ & Number of pulls - arm $i,$ attribute $j$
        \\\hline
        $T_{i}(t)$ & Number of pulls of arm $i$
        \\\hline
        $\hat{\mu}_{ij}(t)$ & Empirical mean reward of arm $i,$ attribute $j$ \\
        \hline
        $\hat{\mu}_i(t)$ & Empirical mean reward of arm $i$ 
        \\\hline
        $\alpha(t)$ & Confidence radii \\
        \hline
        $L_{ij}(t), U_{ij}(t)$ & Attribute confidence bounds \\
        \hline
        $L_{i}(t), U_{i}(t)$ & Arm confidence bounds \\
        \hline
        $\mathcal{F}_{Pt}^A$ & Perfectly feasible attribute set \\
        \hline
        $\partial\mathcal{F}_t^A$ & Almost feasible attribute set \\
        \hline
        ${\mathcal{F}_t^A}$ & Feasible attribute set \\
        \hline
        $\mathcal{F}_{Pt}$ & Perfectly feasible arm set \\
        \hline
        $\partial\mathcal{F}_t$ & Almost feasible arm set \\
        \hline
        ${\mathcal{F}_t}$ & Feasible arm set \\
        \hline
        $\mathcal{P}_t$ & Potential set \\
        \hline
        $\hat{f}$ & Empirical feasibility flag \\
        \hline
        $i_t$ & Best arm from perfect feasible set \\
        \hline
        $i^{\star}_t$ & Best arm from feasible set\\
        \hline

    \end{tabular}
    
\end{table}


\begin{algorithm}[htbp]\label{alg: cap}
\caption{Confidence Set Sampling (CSS-LUCB)}\label{alg:cap}
\begin{algorithmic}[1]
\State $t \gets 1,$ $T_{ij}(t) \gets 1 \ \forall \ (i,j) \in [N] \times [M],$ $T_i(t) \gets 1 \ \forall \ i$
\State Initialize $\partial\mathcal{F}_t^A$, ${\mathcal{F}_t^A}$, $\mathcal{F}_{Pt}$, $\partial\mathcal{F}_t$, ${\mathcal{F}_t}, \mathcal{P}_t$ as $\phi$
\State Sample each of the $N$ arms once
\State Set $\mathcal{F}_{M \times N} = [N]$
\For {time steps $t > M \times N$}
\State Calculate $\hat{\mu}_{ij}$, $\hat{\mu}_i$ $\forall i, j$ using \eqref{eq: emp_mean}
\State Calculate confidence bounds using \eqref{eq: bound_attr} and \eqref{eq: bound_arm}
\State Update $\partial\mathcal{F}_t^A$, ${\mathcal{F}_t^A}$, $\mathcal{F}_{Pt}$, $\partial\mathcal{F}_t$, ${\mathcal{F}_t}$ according to \ref{sec: algo}
\State Find $i^{\star}_t\coloneqq\arg\max\{\hat{\mu}_i(t): i \in \mathcal{F}_{Pt}\}$
\State Update $\mathcal{P}_t$  according to \eqref{eq: potential set}
\State Set $i_t\coloneqq\arg\max\{\hat{\mu}_i(t): i \in \mathcal{F}_{t}\}$
\If {$\mathcal{F}_t\cap \mathcal{P}_t = \phi$}
    \If {$\mathcal{F}_t \neq \phi$}, set $i_{out} = i_t$, $\hat{f}=1$
    \Else{} Set $\hat{f}=0$
    \EndIf
    \State \textbf{break}
\EndIf
\If{$|\mathcal{F}_t|=1$}
    \State Pull $(i_t, j)$ such that $(i_t, j) \in \mathcal{F}_t^A$
    
\Else
    \State Find $i_t$ and $c_t$
    \State Pull $(i, j)$ such that $i \in \{i_t, c_t\}$, $(i, j) \in \mathcal{F}_t^A$
\EndIf
\EndFor
\end{algorithmic}
\end{algorithm}

\subsection{Sampling Criteria}
We consider two cases. The first case is when a single arm is in the \emph{Feasible Set.} In this case, we find $i_t$ defined as the arm with the highest empirical mean reward from the \emph{Feasible Set.} i.e., $i_t\coloneqq\arg\max\{\hat{\mu}_i(t): i \in \mathcal{F}_{t}\}.$

We then check the arm-attribute pairs in the \emph{Feasible Attribute Set} and pull all the attributes of arm $i_t$ in this set. In the second case, i.e., when there is more than one arm in the \emph{Feasible Set,} $i_t$ is defined as the best arm from the \emph{Feasible Set,} and $c_t$ is the \emph{potentially competitor arm}, that is an arm other than $i_t$ from the \emph{Feasible Set,} which has the highest UCB. It is defined as follows:
\begin{equation*}
    c_t\coloneqq\arg\max\{U_i(t): i \in \mathcal{F}_{t}, i \neq i_t\}.
\end{equation*}

We then check the \emph{Feasible Attribute Set} and pull all the arm-attribute pairs corresponding to arms $i_t$ and $c_t$. 
\subsection{Stopping Criteria}
The algorithm stops when there are no competitor arms to be pulled, that is, the set ${\mathcal{F}}_t \cap \mathcal{P}_t = \phi$. We then check the feasible set; if ${\mathcal{F}}_t = \phi$, then the given instance is declared to be infeasible, and the feasibility flag $\hat{f}$ is set to 0. Otherwise, the feasibility flag is set to 1, and the arm $i_t$ is declared the best feasible arm. The pseudo-code of the above algorithm is given in Alg \ref{alg:cap}.

\section{Analytical Performance Guarantees}
In this section, we provide analytical guarantees for the performance of our algorithm. We define the following sets based on the ground truth:
\begin{enumerate}

    \item \emph{Sub-optimal Set:} the set of arms with   mean reward less than that of the best feasible arm. Formally, 
        \begin{equation*}
            \mathcal{S} \coloneqq \begin{cases} 
                \{ i \in [N] : \mu_i < \mu_{i^{\star}}\} & \mathcal{F} \neq \phi \\
                \phi & \mathcal{F} = \phi.
            \end{cases}
    \end{equation*}
    \item \emph{Risky Set:} the set of arms with mean reward more than the best feasible arm. Note that by definition, all these arms are infeasible. Formally, 
    \begin{equation*}
        \mathcal{R}\coloneqq [N]\setminus \{ \mathcal{S} \cup \{i^\star\} \}.
        \label{eq: calc_delta_i_star}
    \end{equation*}
\end{enumerate}
We define $i^{\star\star} \coloneqq \arg\max\limits_{i \in \mathcal{S}} \mu_i $, and for all $i \in \mathcal{S} \setminus \{i^\star\},$ $ \Delta_i \coloneqq |\mu_{i^{\star}} - \mu_i|.$ Further,
\begin{equation}
    \Delta_{i^{\star}} \coloneqq |\mu_{i^{\star}} - \mu_{i^{\star\star}}|.
    \label{eq: calc_delta_i_star}
\end{equation}
Similarly, $\Delta_{ij}^{\text{attr}}\coloneqq |\mu_{ij} - \mu_{\text{TH}}|$. We also define $\Delta_{i}^{\text{attr}}\coloneqq |\min \limits_{j}{\mu_{ij} - \mu_{\text{TH}}}|$.

Further, the separator $\bar{\mu}$ is defined as follows:
\begin{equation}
    \bar{\mu} \coloneqq \begin{cases} 
                                    (\mu_{i^{\star}} + \mu_{i^{\star\star}})/2 & \mathcal{F} \neq \phi \: \text{and } \:\mathcal{S} \neq \phi \\
                                    -\infty & \text{otherwise}. 
                                    \end{cases}
\end{equation}

Next, we define the hardness index of a problem instance, denoted by $H_{\text{id}}$ as
\begin{align*}
        H_{\text{id}} \coloneqq& \frac{1}{\min\{\frac{\Delta_{i^{\star}}}{2}, \Delta_{i^{\star}}^{\text{attr}}\}^2} + \sum\limits_{i \in \mathcal{F} \cap \mathcal{S}}\frac{1}{\left(\frac{\Delta_i}{2}\right)^2} \\
        &+ \sum\limits_{i \in \bar{\mathcal{\overline{F}}} \cap \mathcal{R}}\frac{1}{\left(\Delta_i^{\text{attr}}\right)^2} + \sum\limits_{i \in \bar{\mathcal{\overline{F}}} \cap \mathcal{S}} \frac{1}{\max\{\frac{\Delta_{i}}{2}, \Delta_{i}^{\text{attr}}\}^2}.
    \end{align*}
We also have Empirically \emph{Sub-optimal Set}, \emph{Risky Set}, and \emph{Neutral Set} depending on the separator value and are defined according to the equations given below:
 \begin{align*}\label{eq: subopt}
 \mathcal{S}_t &\coloneqq\{i:U_i(t)<\bar{\mu}\}\\
 \mathcal{R}_t &\coloneqq\{i:L_i(t)>\bar{\mu}\}\\
 \mathcal{N}_t &\coloneqq [N]\setminus(\mathcal{S}_t\cup\mathcal{R}_t)=\{i:L_i(t)\leq\bar{\mu}\leq U_i(t)\}.
 \end{align*}

We now state an upper bound on the number of samples required by CSS-LUCB in the following theorem.
\begin{theorem}[Upper bound]\label{th: upper bound}
    Given an instance and a confidence parameter $ \delta$, with probability at least $1-\delta$, the  CSS-LUCB algorithm succeeds and terminates in $\mathcal{O}\left(H_{\text{id}}\ln \frac{H_{\text{id}}}{\delta}\right)$ samples.
\end{theorem}
The proof is in the appendix below. This result combined with Theorem \ref{th: lower bound} show that CSS-LUCB has near-optimal (up to logarithmic factors) performance with respect to the harness index $H_{\text{id}}$. 

\color{black}
\section{Numerical Results}\label{sec: experiments}

In this section, we present our numerical results. We compare the performance of our policy with two suitably adapted variants of the widely studied action elimination algorithm \cite{cbb-algos}. More specifically, along with the usual elimination of arms whose mean reward is low, we also eliminate arms that have one or more attributes whose UCB is less than the given threshold ($\mu_{\text{TH}}$). In addition, we also simulate an algorithm that divides the problem into two sub-tasks. The first task is to eliminate arms that are infeasible, followed by the second task, which is to identify the best arm in the set of feasible arms. We use action elimination for the second task\color{black}. We refer to this approach as \emph{Feasibility then BAI}.

\begin{table}[!hbt]
		\begin{center}
		\caption{Experimental Data}
		\begin{tabular}{|c|c|c|c|c|c|c|}
   \multicolumn{7}{c}{Experiment 1: $x$ varies from 0.31 to 0.35, $\mu_{\text{TH}} = 0.3$} \\
   \hline
    \rowcolor[HTML]{C0C0C0}			
   \textbf{Arm} & \textbf{Attr1}           & \textbf{Attr2} & \textbf{Attr3} & \textbf{Attr4} & \textbf{Attr5} & \textbf{Mean}  \\ \hline

\textbf{1}   & \textbf{x} & 0.6            & 0.7            & 0.6            & 0.7            & \textbf{0.58 - 0.59} \\ \hline \rowcolor[HTML]{E0E0E0} 
\textbf{2}   & 0.4                      & 0.2            & 0.4            & 0.4            & 0.45           & \textbf{0.37}  \\ \hline \rowcolor[HTML]{E0E0E0} 
\textbf{3}   & 0.15                     & 0.7            & 0.8            & 0.9            & 0.9            & \textbf{0.69}  \\ \hline
\textbf{4}   & \textbf{x}                     & 0.35           & 0.35           & 0.35           & 0.4            & \textbf{0.35 - 0.36} \\ \hline
\textbf{5}   & \textbf{x}                     & 0.35           & 0.35           & 0.35           & 0.35           & \textbf{0.34 - 0.35} \\ \hline

 \multicolumn{7}{c}{\color{white}Experiment 2} \\
   \multicolumn{7}{c}{Experiment 2: $x$ varies from 0.5 to 0.7, $\mu_{\text{TH}} = 0.3$} \\
   \hline
    \rowcolor[HTML]{C0C0C0}			
   \textbf{Arm} & \textbf{Attr1}             & \textbf{Attr2}             & \textbf{Attr3} & \textbf{Attr4} & \textbf{Attr5} & \textbf{Mean} \\ \hline
\textbf{1}   &  0.7 & 0.6                        & 0.8            & 0.7            & 0.9            & \textbf{0.74} \\ \hline
\textbf{2}   & 0.7                        & \textbf{x} & 0.7            & 0.7            & 0.8            & \textbf{0.68 - 0.72} \\ \hline \rowcolor[HTML]{E0E0E0} 
\textbf{3}   & 0.15                       & 0.7                        & 0.8            & 0.9            & 0.9            & \textbf{0.69} \\ \hline \rowcolor[HTML]{E0E0E0} 
\textbf{4}   & 0.15                       & 0.9                        & 0.9            & 0.9            & 0.8            & \textbf{0.73} \\ \hline \rowcolor[HTML]{E0E0E0} 
\textbf{5}   & 0.1                        & 0.9                        & 0.9            & 0.8            & 0.8            & \textbf{0.7} \\ \hline 

 \multicolumn{7}{c}{\color{white}Experiment 3} \\
   \multicolumn{7}{c}{Experiment 3: $x$ varies from 0.05 to 0.3, $\mu_{\text{TH}} = 0.35$} \\
   \hline
			\rowcolor[HTML]{C0C0C0} 
\textbf{Arm}                       & \textbf{Attr1}               & \textbf{Attr2}             & \textbf{Attr3} & \textbf{Attr4} & \textbf{Attr5} & \textbf{Mean}                         \\ \hline
\textbf{1}                         &  0.5   & 0.6                        & 0.6            & 0.5            & 0.8            & \textbf{0.6}                          \\ \hline
\textbf{2} & 0.7                          & 0.5 & 0.4            & 0.4            & 0.6            & \textbf{0.52} \\ \hline \rowcolor[HTML]{E0E0E0} 
\textbf{3} & \textbf{x} & 0.5                        & 0.9            & 0.8            & 0.9            & \textbf{0.63 - 0.68} \\ \hline \rowcolor[HTML]{E0E0E0} 
\textbf{4} & 0.6                          & 0.2                        & 0.4            & 0.7            & 0.6            & \textbf{0.5}  \\ \hline \rowcolor[HTML]{E0E0E0} 
\textbf{5} & 0.3                          & 0.7                        & 0.4            & 0.9            & 0.5            & \textbf{0.56} \\ \hline
		\end{tabular}
        \label{tab:experiment-params-1}
		\end{center}
\end{table}
 
For the results presented in this section, the reward of each attribute of each arm is an independent stochastic process with the beta distribution. We perform three different experiments:
\begin{enumerate}
    \item We vary the mean reward of an attribute of the best arm, making it more difficult to assess the feasibility of that arm.       
    \item We vary the mean reward of a suboptimal arm, which increases the hardness of detecting the arm with highest mean.     
    \item The arm with the highest mean is infeasible (\emph{Risky arm).} We vary the mean reward of one of its attributes, making it difficult to detect its in-feasibility.      
\end{enumerate}

We set $\delta = 0.1$, $N = 5,$ and $M = 5.$ The values of various parameters are given in Table \ref{tab:experiment-params-1}. (The shaded rows depict infeasible arms.)

\begin{figure}[htbp]
\begin{subfigure}{0.5\textwidth}
    \includegraphics[width=\textwidth]{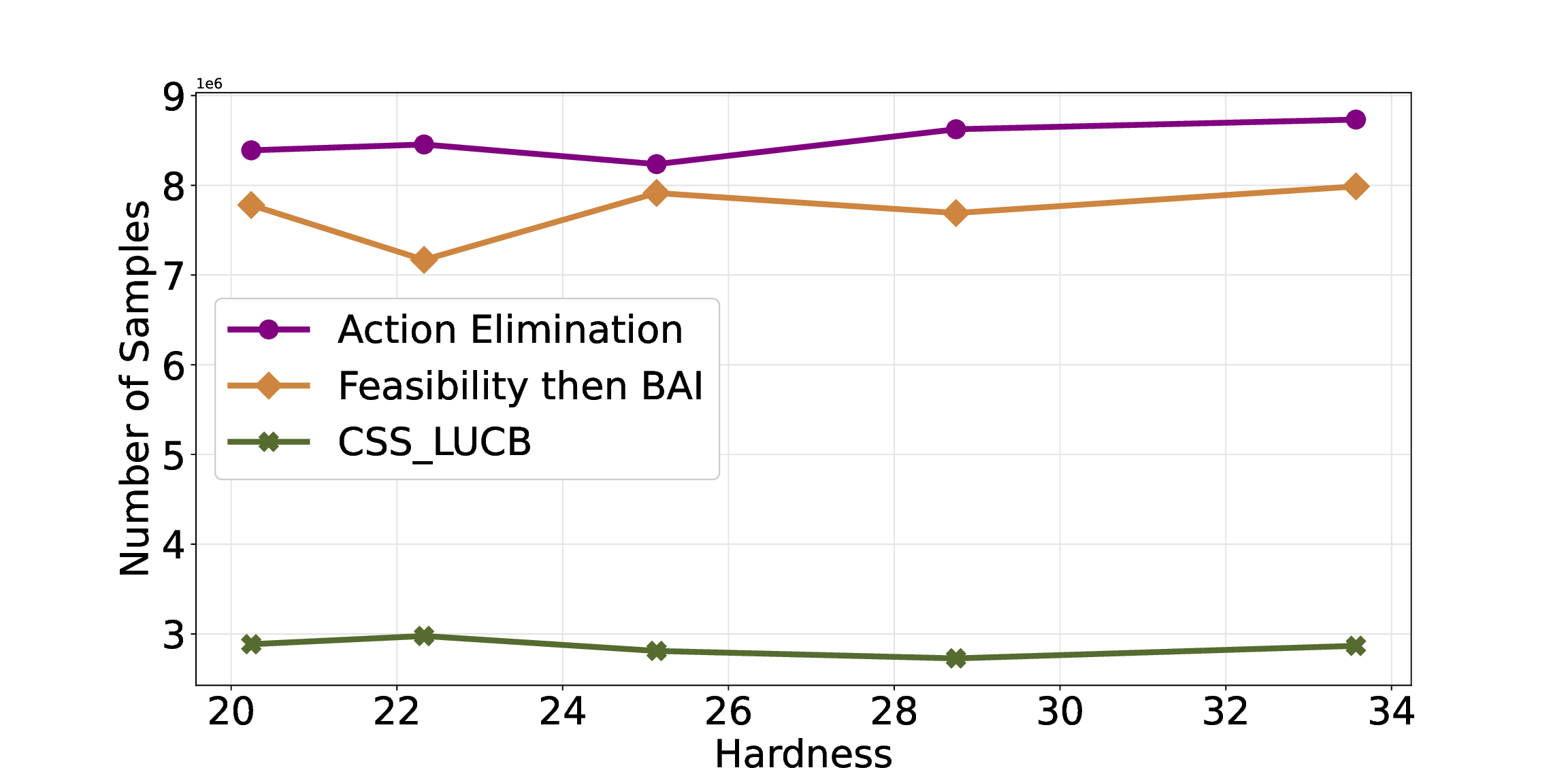}
    \subcaption{Experiment 1}
    \label{fig:first}
\end{subfigure}
\begin{subfigure}{0.5\textwidth}
    \centering
    \includegraphics[width=\textwidth]{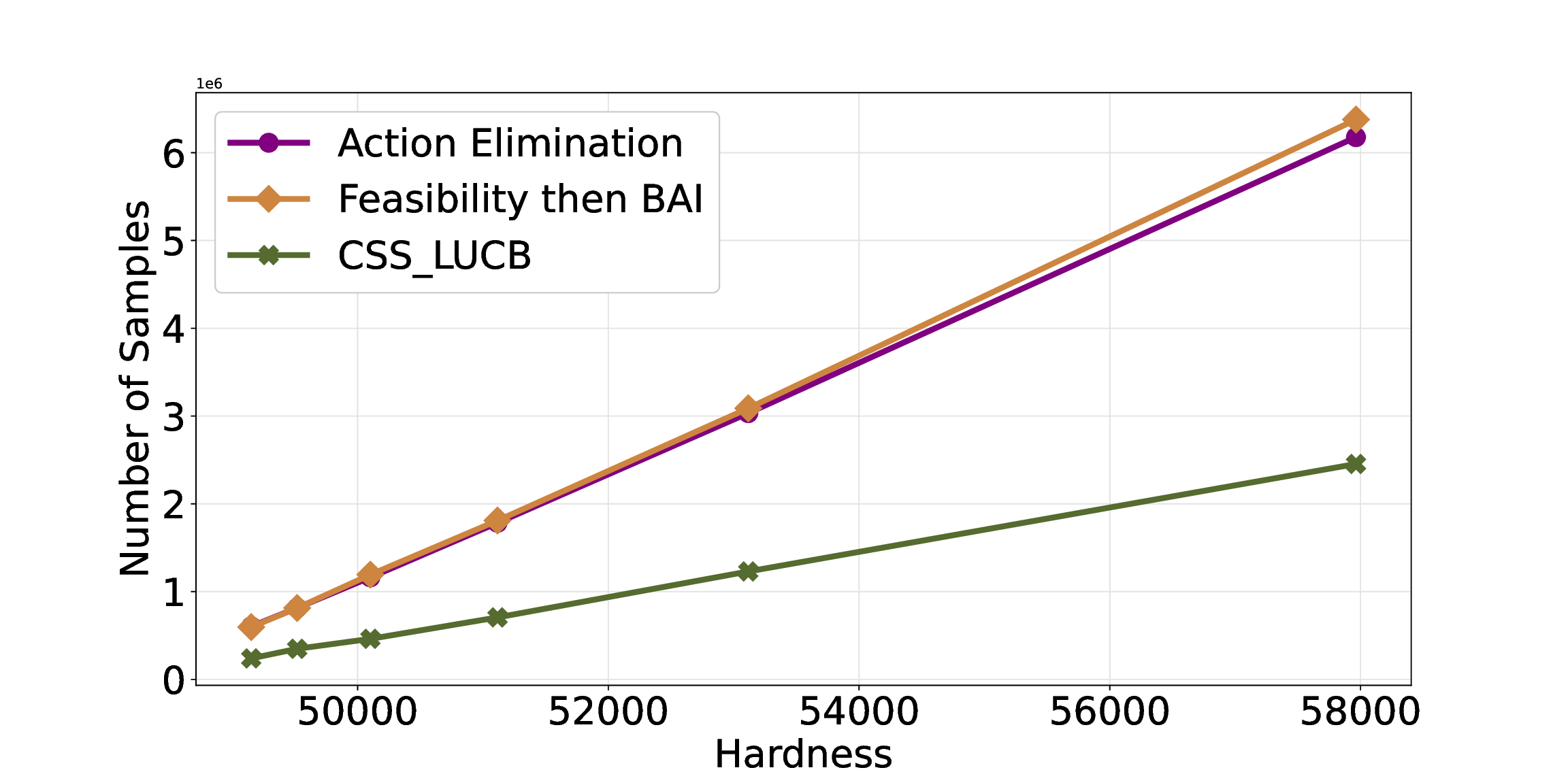}
    \caption{Experiment 2}
    \label{fig:second}
\end{subfigure}
\begin{subfigure}{0.5\textwidth}
    \centering
    \includegraphics[width=\textwidth]{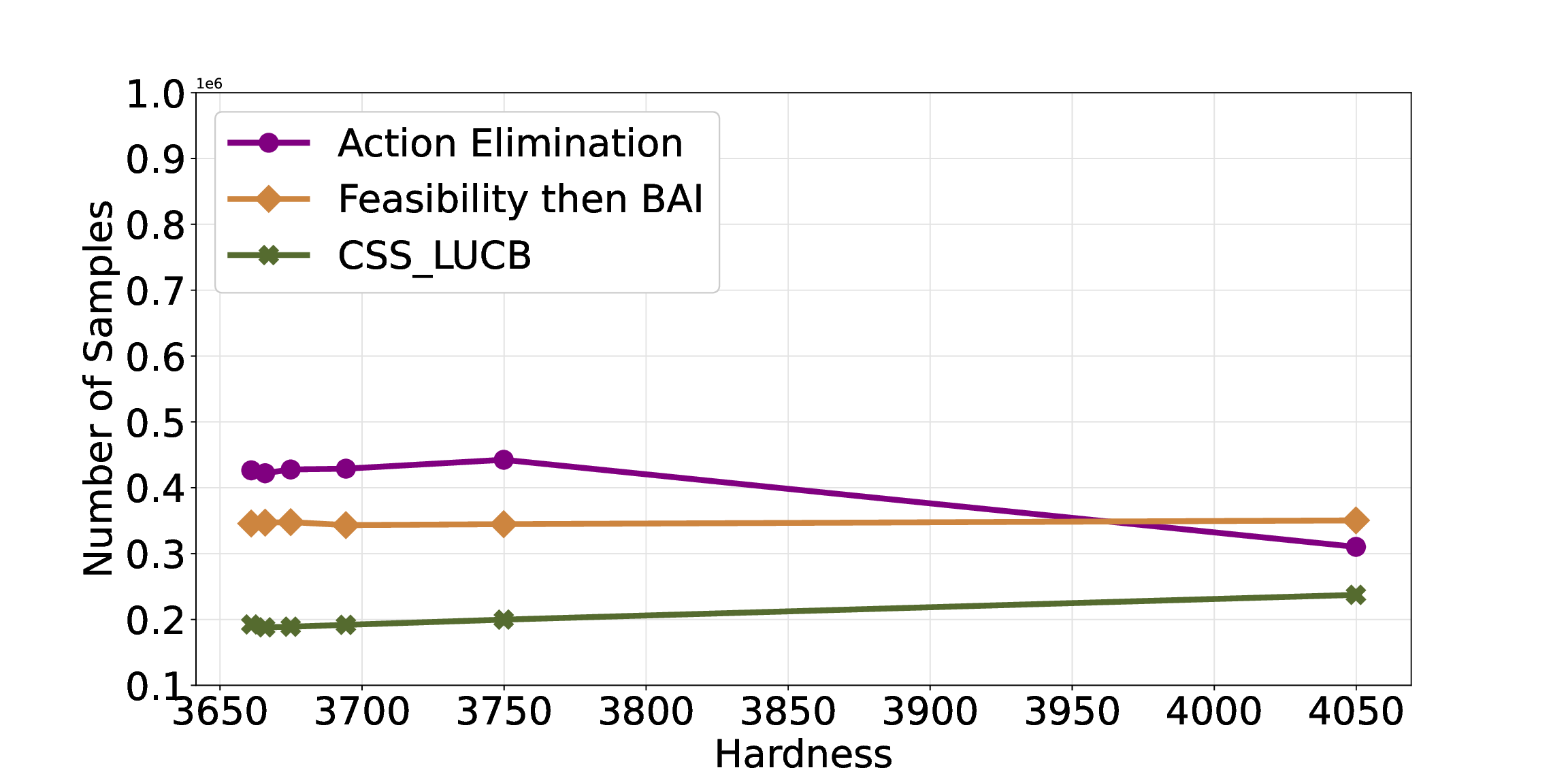}
    \caption{Experiment 3}
    \label{fig:third}
\end{subfigure}       
\caption{Sample-complexity as a function of the hardness}
\label{fig: compare_algo}
\end{figure}

The number of samples required by each algorithm is plotted against the hardness index, $H_{\text{id}}$. The results are shown in Fig.\ref{fig: compare_algo}. Our algorithm outperforms the standard algorithms in all the experiments\color{black}.
\color{black}

\begin{figure}[htbp]
\begin{subfigure}{0.48\textwidth}
    \includegraphics[width=\textwidth]{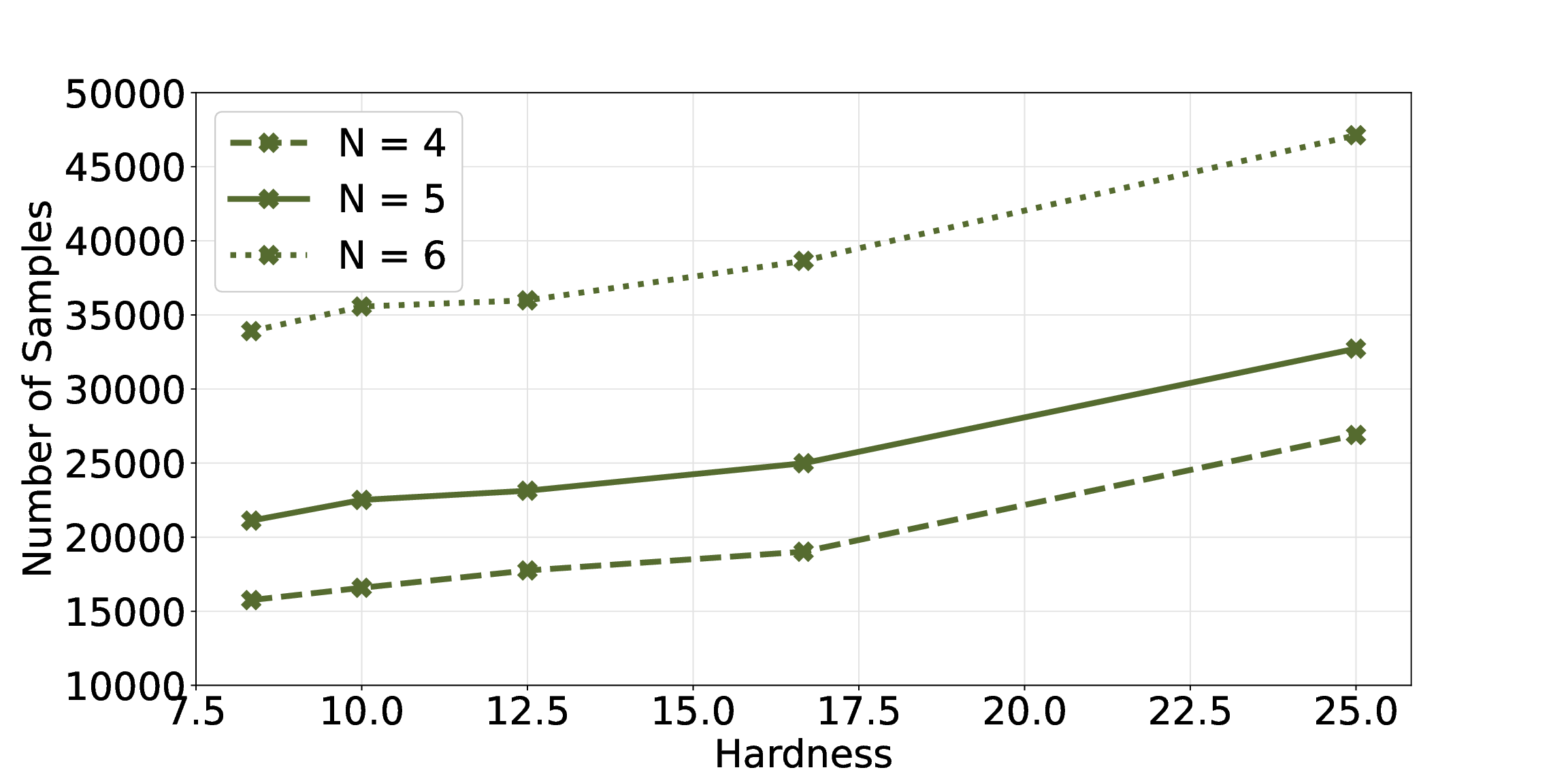}
    \subcaption{Varying number of arms}
    \label{fig: vary_N}
\end{subfigure}
 \begin{subfigure}{0.48\textwidth}
     \centering
     \includegraphics[width=\textwidth]{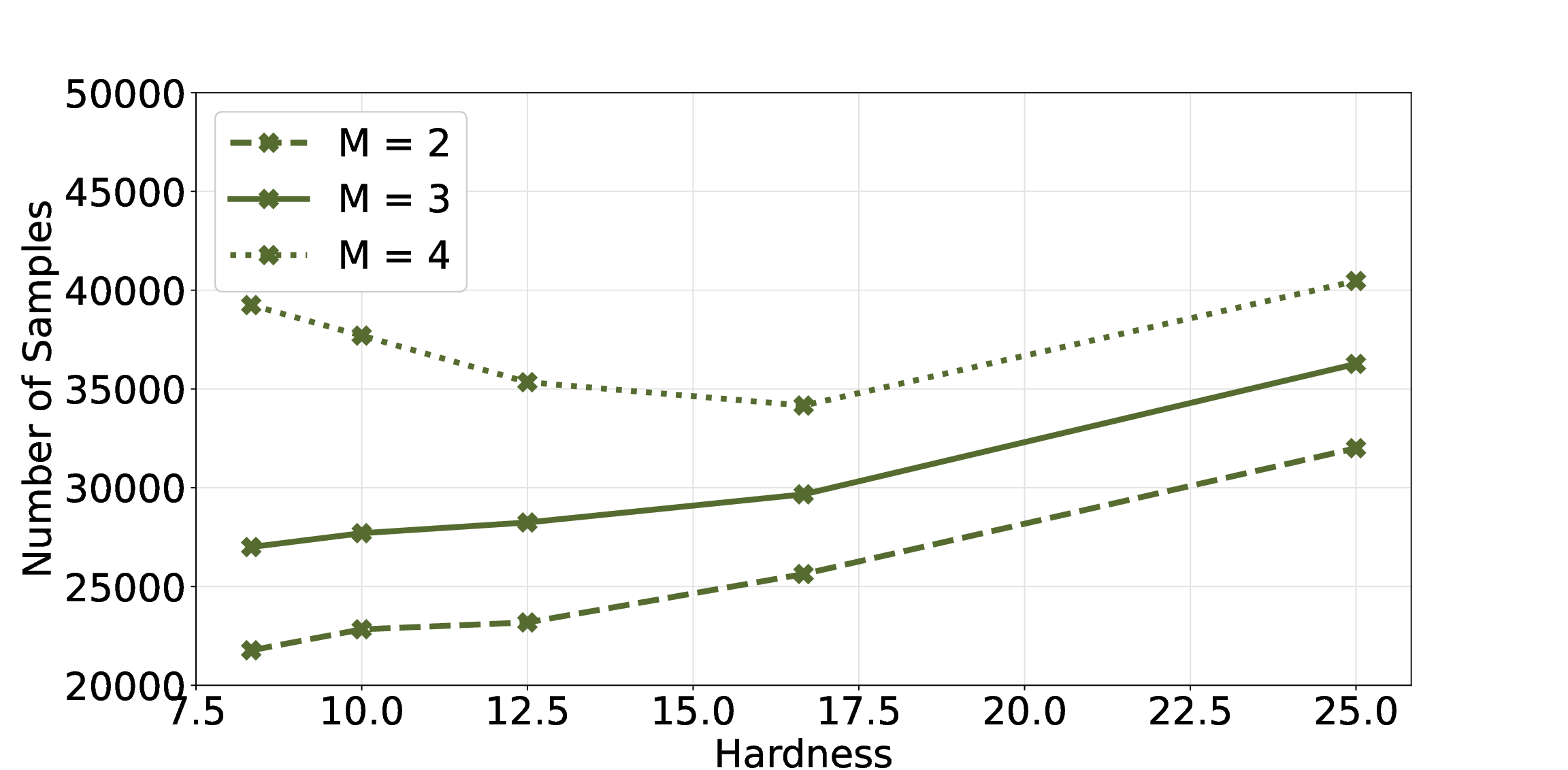}
     \caption{Varying number of attributes}
     \label{fig: vary_M}
 \end{subfigure}     
\caption{Sample complexity of CSS-LUCB on varying the number of arms and attributes}
\label{fig: compare_arm_attr}
\end{figure}

Next, we vary the number of arms and attributes, comparing the results of the CSS-LUCB algorithm in the case where the arm with the highest mean was infeasible, and the results are shown in Fig. \ref{fig: compare_arm_attr}. In Fig \ref{fig: vary_N}, we plot the sample complexity as a function of the number of arms keeping the number of attributes constant. For $N \in \{ 4, 5, 6\},$ we consider arms $1, 2, \cdots, N,$ shown in Table \ref{tab:experiment-params-2}. In Fig \ref{fig: vary_M}, we vary the number of attributes for a fixed number of arms. For $M \in \{2, 3, 4\},$ we consider the first two, three, and four attributes, respectively, from Table \ref{tab:experiment-params-2}. As expected,
sample complexity increases as the number of
arms or attributes increases.

    

\begin{table}[!hbt]
		\begin{center}
		\caption{Data for varying instance parameters}
		\begin{tabular}{|c|c|c|c|}
   \multicolumn{4}{c}{Varying $N$, fixed $M=2$, $\mu_{\text{TH}} = 0.5$, x varies from 0.38 to 0.46} \\
   \hline \rowcolor[HTML]{C0C0C0} 
        \textbf{Arm} & \textbf{Attribute 1} & \textbf{Attribute 2} & \textbf{Mean} \\
        \hline
        1 & 0.6 & 0.7 & \textbf{0.65}\\
        \hline \rowcolor[HTML]{E0E0E0} 
        2 & \textbf{x} & 0.9 & \textbf{0.64 - 0.68}\\
        \hline \rowcolor[HTML]{E0E0E0} 
        3 & 0.3 & 0.55 & \textbf{0.425}\\
        \hline
        4 & 0.55 & 0.55 & \textbf{0.55} \\
        \hline \rowcolor[HTML]{E0E0E0} 
        5 & 0.2 & 0.4 & \textbf{0.3}\\
        \hline
        6 & 0.55 & 0.6 & \textbf{0.575}\\
        \hline
    \end{tabular}
    \begin{tabular}{|c|c|c|c|c|c|}
 \multicolumn{6}{c}{\color{white}Experiment 2} \\
   \multicolumn{6}{c}{Varying $M$, fixed $N=5$, $\mu_{\text{TH}}=0.5$, x varies from 0.38 to 0.46} \\
   \hline \rowcolor[HTML]{C0C0C0} 
   \textbf{Arm} & \textbf{Attr 1} & \textbf{Attr 2} & \textbf{Attr 3} & \textbf{Attr 4} & \textbf{Mean}\\
        \hline
        1 & 0.55 & 0.6 & 0.65 & 0.7 & \textbf{0.625}\\
        \hline \rowcolor[HTML]{E0E0E0} 
        2 & \textbf{x} & 0.9 & 0.7 & 0.8 & \textbf{0.695 - 0.715}\\
        \hline \rowcolor[HTML]{E0E0E0} 
        3 & 0.3 & 0.55 & 0.4 & 0.6 & \textbf{0.463}\\
        \hline
        4 & 0.55 & 0.55 & 0.55 & 0.55 & \textbf{0.55}\\
        \hline \rowcolor[HTML]{E0E0E0} 
        5 & 0.2 & 0.4 & 0.3 & 0.55 & \textbf{0.363}\\
        \hline
		\end{tabular}
        \label{tab:experiment-params-2}
		\end{center}
\end{table}

\newpage
\onecolumn

\begin{center}
APPENDIX
\end{center}
\begin{center}
PROOF OF LOWER BOUND
\end{center}

\begin{lem}[Reverse Pinsker's inequality \cite{Friedrich2019}]
	\label{thm:pinsker}

	{Let $P$ and $Q$ be two distributions that are defined in the same finite space 
		$\mathcal{A}$ and have 
		the same support.} We have
	\begin{align*}
		\frac{1}{2} \mathrm{KL}(P,Q)
		\le \frac{1}{  \alpha_Q} \delta(P,Q)^2
	\end{align*}
	where
	$
	\delta(P,Q) := \sup\{| P(A)-Q(A)| \,:\, A \subset \mathcal{A}   \} 
	=\frac{1}{2} \sum_{x\in  \mathcal{A}} |P(x) - Q(x)|$
	is the total variational distance,
	and  
	$\alpha_Q := \min_{x\in \mathcal{A}: Q(x)>0} Q(x)
	$.
\end{lem}

\noindent We fix a $\delta$-PAC algorithm and assume that all attributes of all the arms have Bernoulli reward distributions with expected reward of the $j^{th}$ attribute of the $i^{th}$ arm given by $\mu_{ij}$. Thus, the reward distribution of an arm-attribute pair $(\nu_{ij})$ is given by: 
\begin{equation*}
\nu_{ij} =
\begin{cases} 
1, & \text{w.p. } \mu_{ij}, \\ 
0, & \text{w.p } 1 - \mu_{ij}.
\end{cases}
\end{equation*}
We first look at two simpler cases before moving to the generic case with all types of arms: \\
(i) All arms are infeasible : ($\exists$ $j$ such that $\mu_{ij} < \mu_{\text{TH}}$ $\forall i \in N$) \\
(ii) All arms are feasible : ($\mu_{ij} \geq \mu_{\text{TH}}, \forall i \in N, \forall j \in M$) \\
Subsequently, we analyze each of the cases separately. In each case, we construct $N + 1$  environments with the expected reward of the $j^{th}$ attribute of the $i^{th}$ arm in environment $k$ ($\mu_{ij}^{(k)}$) defined as follows : 
\begin{equation*}
\mu_{ij}^{(k)} =
\begin{cases} 
\mu_{ij}', & \text{if } i = k, \\ 
\mu_{ij}, & \text{otherwise}.
\end{cases}
\end{equation*}
where $\mu_{ij}'$ will be defined in each case. Thus, we keep everything  unchanged in instance 0, and in instance $k$, we only modify the expected rewards of attributes of the $k^{th}$ arm. Further we define, $\mathcal{G}_r^{(k)}$ as the sequence of pulled arms and observed rewards upto and including round $r$ in environment $k$. Assuming that the stopping time in instance $k$ is $\tau_{(k)}$, we denote $\mathcal{G}_{\tau_{(k)}}^{(k)}$ by $\mathcal{G}_k$ for simplicity.\\

\noindent Case (i) : 
In this case, as there is no feasible arm, the algorithm should output $i_{out} = \phi$ under instance 0. Let $\mathcal{S}$ be the set of all attributes of an arm, i.e. $S = \{1,2,...,M\} $ and $\mathcal{S}_k$ be the set of all attributes of arm $k$ for which $ \mu_{kj} < \mu_{\text{TH}}$. We define the other instances as follows : 
\begin{equation*}
\mu_{kj}' =
\begin{cases} 
\mu_{\text{TH}} + \epsilon , & \text{if } j \in \mathcal{S}_
k \\ 
\mu_{kj}, & \text{otherwise}.
\end{cases}
\end{equation*}
Thus, we ensure that, under instance $k$, arm $k$ is the unique optimal feasible arm as $\mu_{kj}^{(k)} \geq \mu_{\text{TH}} \ (\forall j \in M)$, while all the other arms remain infeasible. Since the algorithm $\pi$ is $\delta$-PAC, we have $\mathbb{P}_{\mathcal{G}_0}[i_{out} = k] < \delta$, and for all other instances $1 \leq k \leq N$, we have $\mathbb{P}_{\mathcal{G}_k}[i_{out} \neq k] < \delta$ or in other words, $\mathbb{P}_{\mathcal{G}_k}[i_{out} = k] \geq 1 - \delta$.

\begin{lem}[Lemma 1 in \cite{Kaufmann2016}] \label{lemma:lb_KLdecomp}
	For any $1\le j \le N$,
	{\setlength\abovedisplayskip{.5em}
		\setlength\belowdisplayskip{.4em}
		\begin{align*}
			& 
			\sum_{i=1}^N \mathbb{E}_{ \mathcal{G}_0} [T_{i}(\tau^{(0)})] \cdot d\big( \mu_i^{(0)}, \mu_i^{(k)} \big)
			\ge 
			\sup_{\mathcal{E} \in \mathcal{G}_0  } 
			d\big(    \mathbb{P}_{\mathcal{G}_0} (\mathcal{E})  ,~   \mathbb{P}_{\mathcal{G}_k} (\mathcal{E})   \big).
			\nonumber 
	\end{align*}}     
\end{lem} 

\noindent In our case, as $\mu_i^{(0)} = \mu_i^{(k)} \ (\forall i \neq k)$, the inequality simplies into the following :
		\begin{align*}
			& 
		   \mathbb{E}_{ \mathcal{G}_0} [T_{k}(\tau^{(0)})]
			\ge \frac{
			\sup_{\mathcal{E} \in \mathcal{G}_0  } 
			d\big(    \mathbb{P}_{\mathcal{G}_0} (\mathcal{E})  ,~   \mathbb{P}_{\mathcal{G}_k} (\mathcal{E})   \big).}{d\big( \mu_k^{(0)}, \mu_k^{(k)} \big)}
			\nonumber 
	\end{align*}
Now we use Lemma \ref{lemma:lb_KLdecomp} to lower bound $\mathbb{E}_{\mathcal{G}_0} [T_i( \tau^{(0)}  )]$ with the KL divergence. We define event $\mathcal{E}_k$ to be \{$i_{out} = k$\} and using the arguments made above, we get:
		\begin{align*}
			& 
		   \mathbb{E}_{ \mathcal{G}_0} [T_{k}(\tau^{(0)})]
			\ge \frac{d(\delta, 1 - \delta)}{d\big( \mu_k^{(0)}, \mu_k^{(k)} \big)}
			\nonumber 
\end{align*}
Therefore, we have
\begin{align*}
    \mathbb{E}_{ \mathcal{G}_0} [\tau^{(0)}] & \geq \sum_{k=1}^{N}  \mathbb{E}_{ \mathcal{G}_0} [T_{k}(\tau^{(0)})]\\ & \geq \sum_{k=1}^{N} \frac{d(\delta, 1 - \delta)}{d\big( \mu_k^{(0)}, \mu_k^{(k)} \big)}
			\nonumber \\ &  \overset{\text{(a)}}{\geq} \text{ln}\bigg(\frac{1}{2.4\delta}\bigg) \sum_{k=1}^{N} \frac{1}{{d\big( \mu_k^{(0)}, \mu_k^{(k)} \big)}} \\ & \geq \text{ln}\bigg(\frac{1}{2.4\delta}\bigg) \sum_{k=1}^{N} \frac{1}{{d\bigg( \frac{\sum_{j \in S} \mu_{kj}}{M},  \frac{\sum_{j \in S} \mu_{kj}'}{M} \bigg)}} \\ & \geq \text{ln}\bigg(\frac{1}{2.4\delta}\bigg) \sum_{k=1}^{N} \frac{1 }{{d\bigg( \frac{\sum_{j \in S} \mu_{kj}}{M},  \frac{\sum_{j \in S\setminus S_k} \mu_{kj} + \sum_{j \in S_k} \mu_{\text{TH}} + \epsilon}{M} \bigg)}} \\ &  \overset{\text{(b)}}{\geq} \text{ln}\bigg(\frac{1}{2.4\delta}\bigg) \sum_{k=1}^{N} \frac{1}{{d\bigg( \frac{\sum_{j \in S} \mu_{kj}}{M},  \frac{\sum_{j \in S\setminus S_k} \mu_{kj}}{M}  + \frac{|S_k| \cdot \mu_{\text{TH}}}{M}\bigg)}} \\ & \overset{\text{(c)}}{\geq} \text{ln}\bigg(\frac{1}{2.4\delta}\bigg)\sum_{k=1}^{N} \frac{c}{{2\bigg( \frac{\sum_{j \in S} \mu_{kj}}{M} - \bigg(\frac{\sum_{j \in S\setminus S_k} \mu_{kj}}{M}  + \frac{|S_k| \cdot \mu_{\text{TH}}}{M}\bigg)\bigg)}^2} \\ & \geq \text{ln}\bigg(\frac{1}{2.4\delta}\bigg)\sum_{k=1}^{N} \frac{1}{\bigg(\sum_{j \in S_k}\big(\mu_{\text{TH}} - \mu_{kj} \big)\bigg)^2} \cdot \frac{M^2c}{2} \\ & \geq \text{ln}\bigg(\frac{1}{2.4\delta}\bigg) \sum_{k=1}^{N} \frac{1}{|S_k|^2\big(\mu_{\text{TH}} - \min_{j} \mu_{kj} \big)^2} \cdot \frac{M^2c}{2} \\ & \geq \text{ln}\bigg(\frac{1}{2.4\delta}\bigg) \frac{M^2c}{2\cdot\max_{k}|S_k|^2} \sum_{k=1}^{N}\frac{1}{\big(\Delta_{k}^{attr} \big)^2} \\ & \geq \text{ln}\bigg(\frac{1}{2.4\delta}\bigg) \frac{M^2c\ H_{\text{id}}}{2\cdot\max_{k}|S_k|^2} \\ &  \overset{\text{(d)}}{\geq} \text{ln}\bigg(\frac{1}{2.4\delta}\bigg) c_1 {H_{\text{id}}}. 
\end{align*}
Here, (a) follows from the fact that the KL Divergence between Bern($\delta$) and Bern($1-\delta$)
can be lower bounded by ln($\frac{1}{2.4\delta}$) and (b) is obtained by taking supremum over all $\epsilon > 0$. Inequality (c) follows from \ref{thm:pinsker} and the constant c is given by: 
\begin{align*}
    c = \min\bigg(\frac{\sum_{j \in S\setminus S_k} \mu_{kj}}{M}  + \frac{|S_k| \cdot \mu_{\text{TH}}}{M}
, 1- \Big(\frac{\sum_{j \in S\setminus S_k} \mu_{kj}}{M}  + \frac{|S_k| \cdot \mu_{\text{TH}}}{M}\Big)\bigg).
\end{align*}

\noindent In (d) the constant $c_1$ depends on the original instance and is given by:

\begin{equation*}
    c_1 = \frac{M^2c}{2\cdot \max_{k}|S_k|^2} \cdot 
\end{equation*}

\noindent Thus, we conclude that in this case the number of samples required for the algorithm to terminate can be lower bounded by $c_1H_\text{id}$ or in other words, the number of samples are of the order of the hardness index ($H_\text{id}$).\\

\noindent Case (ii) : All arms are feasible. Let's assume without loss of generality that arm 1 is the best arm and define the event $\mathcal{E}$ to be $\{i_{out} \neq 1\}$. So, under instance 0, the algorithm outputs arm 1 as the best arm and hence $\mathbb{P}_{\mathcal{G}_0}[i_{out} \neq 1] < \delta$. Let $j'$ be the attribute of arm 1 with the least expected reward i.e. $ j' = {\arg\min_{j}} \mu_{1j} $. Under instance 1, we define $\mu_{1j}'$ as follows:
\begin{equation*}
\mu_{1j}' =
\begin{cases} 
\mu_{\text{TH}} - \epsilon , &  j = j'\\ 
\mu_{1j}, & \text{otherwise}.
\end{cases}
\end{equation*}
By doing this, we make arm 1 infeasible and thus $\mathbb{P}_{\mathcal{G}_1}[i_{out} \neq 1] \geq 1-\delta$. Now, for all the other instances $(2,...,N)$, we define $\mu_{kj}' = \mu_1 + \epsilon \ (\forall j  \in M)$. Thus, $\mu_{k}'$ becomes equal to $\mu_1 + \epsilon$, making arm $k$ the best arm and $\mathbb{P}_{\mathcal{G}_k}[i_{out} \neq 1] \geq 1-\delta$. Again, using the arguments made above and Lemma \ref{lemma:lb_KLdecomp}, we can write:
\begin{align*}
    \mathbb{E}_{ \mathcal{G}_0} [\tau^{(0)}] & \geq \sum_{k=1}^{N}  \mathbb{E}_{ \mathcal{G}_0} [T_{k}(\tau^{(0)})]\\ & \geq \sum_{k=1}^{N} \frac{d(\delta, 1 - \delta)}{d\big( \mu_k^{(0)}, \mu_k^{(k)} \big)} \nonumber \\ & \geq \text{ln}\bigg(\frac{1}{2.4\delta}\bigg) \bigg(\frac{1}{d\big( \mu_1^{(0)}, \mu_1^{(1)} \big)} + \sum_{k=2}^{N}\frac{1}{{d\big( \mu_k^{(0)}, \mu_k^{(k)} \big)}} \bigg) \\ & \geq \text{ln}\bigg(\frac{1}{2.4\delta}\bigg) \bigg(\frac{1}{ d\big (\mu _{1}, \mu_1 - (\frac{\mu_{1j'} + \epsilon - \mu_{\text{TH}}}{M}) \big)} + \sum_{k=2}^{N}\frac{1}{{d\big( \mu_k, \mu_1 + \epsilon \big)}} \bigg) \\ & \geq \text{ln}\bigg(\frac{1}{2.4\delta}\bigg) \bigg(\frac{1}{ d\big (\mu _{1}, \mu_1 - (\frac{\mu_{1j'} - \mu_{\text{TH}}}{M}) \big)} + \sum_{k=2}^{N}\frac{1}{{d\big( \mu_k, \mu_1 \big)}} \bigg) \\ & \geq \text{ln}\bigg(\frac{1}{2.4\delta}\bigg) \bigg(\frac{M^2c}{2 \big(\mu_{1j'} - \mu_{\text{TH}}\big)^2} + \sum_{k=2}^{N} \frac{\min(\mu_1, 1-\mu_1)}{2\big(\mu_{1} - \mu_{k}\big)^2}\bigg) \\ & \geq \text{ln}\bigg(\frac{1}{2.4\delta}\bigg) \bigg(\frac{M^2c}{2(\Delta_{1}^{attr})^2} + \frac{\min(\mu_1, 1-\mu_1)}{16(\Delta_1/2)^2} + \sum_{k=2}^{N} \frac{\min(\mu_1, 1-\mu_1)}{16(\Delta_k/2)^2}\bigg) \\ & \geq \text{ln}\bigg(\frac{1}{2.4\delta}\bigg) \min\bigg( \frac{M^2c}{2} , \frac{\min(\mu_1, 1-\mu_1)}{16}\bigg) \bigg(\frac{1}{(\Delta_1^{attr})^2} + \frac{1}{(\Delta_1/2)^2} + \sum_{k=2}^{N}\frac{1}{(\Delta_k/2)^2}\bigg) \\ & \geq \text{ln}\bigg(\frac{1}{2.4\delta}\bigg) \min\bigg( \frac{M^2c}{2} , \frac{\min(\mu_1, 1-\mu_1)}{16}\bigg) \bigg(\frac{1}{\text{min}\big((\Delta_1^{attr})^2, (\Delta_1/2)^2 \big)} + \sum_{k=2}^{N}\frac{1}{(\Delta_k/2)^2}\bigg) \\ & \geq \text{ln}\bigg(\frac{1}{2.4\delta}\bigg) \min\bigg( \frac{M^2c}{2} , \frac{\min(\mu_1, 1-\mu_1)}{16}\bigg) \cdot H_\text{id} \\ & \geq \text{ln}\bigg(\frac{1}{2.4\delta}\bigg) c_2H_\text{id}.
\end{align*}
Even for this case, we conclude that the number of samples required for an algorithm to terminate are of the order of hardness index. Again, $c$ and $c_2$ both depend on the original instance and are given by:\\

\begin{align*}
    c & =  \min\bigg(\mu_1 - \big(\frac{\mu_{1j'} - \mu_{\text{TH}}}{M}\big), 1 - \Big(\mu_1 - \big(\frac{\mu_{1j'} - \mu_{\text{TH}}}{M}\big)\Big)\bigg) \\
    c_2 & = \min\bigg( \frac{M^2c}{2} , \frac{\min(\mu_1, 1-\mu_1)}{16}\bigg).
\end{align*}

\noindent We now look at the general case where we have all types of arms. We again assume that arm 1 is the best arm and $\mathcal{E} = \{i_{out} \neq 1\}$. Note that for $k$ such that arm $k \in ({\mathcal{F}}\cap\mathcal{S} + i^{*})$, we can create instances similar to case (ii) and obtain a lower bound. Likewise for all arms in $ (\bar{\mathcal{F}}^c\cap\mathcal{R})$, we increase the expected rewards of infeasible attributes similar to case (i), and given this, arm k now becomes the best arm as it was already risky and now we made it feasible as well. Further calculations for the bound are similar to that in case 1. Now, we only need to analyze the arms that belong to $ (\bar{\mathcal{F}}^c\cap\mathcal{S})$. Similar to the feasible and suboptimal arms, we define  $\mu_{kj}' = \mu_1 + \epsilon \ (\forall j  \in M, \forall k \in \bar{\mathcal{F}}^c\cap\mathcal{S})$. For all these instances, arm $k$ is now the best arm, and hence $\mathbb{P}_{\mathcal{G}_k}[i_{out} \neq 1] \geq 1-\delta$. So, we have:
\begin{align*}
     \sum_{k \in \bar{\mathcal{F}}^c\cap\mathcal{S}}  \mathbb{E}_{ \mathcal{G}_0} [T_{k}(\tau^{(0)})] & \geq  \sum_{k \in \bar{\mathcal{F}}^c\cap\mathcal{S}} \frac{d(\delta, 1 - \delta)}{d\big( \mu_k^{(0)}, \mu_k^{(k)} \big)} \nonumber \\ & \geq \sum_{k \in \bar{\mathcal{F}}^c\cap\mathcal{S}} \frac{d(\delta, 1 - \delta)}{d\big( \mu_k, \mu_1 + \epsilon \big)} \nonumber \\ & \geq \text{ln}\bigg(\frac{1}{2.4\delta}\bigg) \sum_{k \in \bar{\mathcal{F}}^c\cap\mathcal{S}} \frac{1}{d\big( \mu_k, \mu_1 \big)} \\ & \geq \text{ln}\bigg(\frac{1}{2.4\delta}\bigg) \sum_{k \in \bar{\mathcal{F}}^c\cap\mathcal{S}} \frac{\min(\mu_1, 1-\mu_1)}{2\big(\mu_1 - \mu_{k}\big)^2} \\ & \geq \text{ln}\bigg(\frac{1}{2.4\delta}\bigg) \sum_{k \in \bar{\mathcal{F}}^c\cap\mathcal{S}} \frac{\min(\mu_1, 1-\mu_1)}{8\cdot\text{max}\big((\Delta_k^{attr})^2, (\Delta_k/2)^2 \big)} \\ & \geq \text{ln}\bigg(\frac{1}{2.4\delta}\bigg) c_3 \sum_{k \in \bar{\mathcal{F}}^c\cap\mathcal{S}} \frac{1}{\text{max}\big((\Delta_k^{attr})^2, (\Delta_k/2)^2 \big)}, 
\end{align*}
where $c_3 = \frac{\min(\mu_1, 1-\mu_1)}{8}$. Putting together the results obtained from the previous cases, we have:
\begin{align*}
     \mathbb{E}_{ \mathcal{G}_0} [\tau^{(0)}] & \geq      \sum_{k \in \bar{\mathcal{F}}^c\cap\mathcal{S}}  \mathbb{E}_{ \mathcal{G}_0} [T_{k}(\tau^{(0)})] +      \sum_{k \in \{ {\mathcal{F}}\cap\mathcal{S} + i^{*}\}}  \mathbb{E}_{ \mathcal{G}_0} [T_{k}(\tau^{(0)})] +      \sum_{k \in \bar{\mathcal{F}}^c\cap\mathcal{R}}  \mathbb{E}_{ \mathcal{G}_0} [T_{k}(\tau^{(0)})] \\ & \geq  \text{ln}\Bigg(\frac{1}{2.4\delta}\Bigg) \Bigg(\sum_{k \in \bar{\mathcal{F}}^c\cap\mathcal{S}} \frac{c_3}{\text{max}\big((\Delta_k^{attr})^2, (\Delta_k/2)^2 \big)} + \frac{c_2}{\text{min}\big((\Delta_1^{attr})^2, (\Delta_1/2)^2 \big)} + \sum_{{\mathcal{F}}\cap\mathcal{S}}\frac{c_2}{(\Delta_k/2)^2} + \sum_{k \in \bar{\mathcal{F}}^c\cap\mathcal{R}}\frac{c_1}{\big(\Delta_k^{attr} \big)^2}\Bigg) \\ & \geq \text{ln}\Bigg(\frac{1}{2.4\delta}\Bigg) \cdot \min(c_1,c_2,c_3) \cdot H_{\text{id}} \\ & = \text{ln}\Bigg(\frac{1}{2.4\delta}\Bigg) \cdot C(\nu, \mu_{\text{TH}}) \cdot H_{\text{id}},
\end{align*}
which completes the proof of the lower bound.\\
\begin{center}
PROOF OF UPPER BOUND
\end{center}
To derive an upper bound on the sample complexity of CSS-LUCB, we first define some empirical sets as follows:
\begin{align*}
	&\mathcal{S}_t :=\{i:U_i^\mu(t)< \bar{\mu} \},\quad 
	\mathcal{R}_t:=\{i:L_i^\mu(t)> \bar{\mu} \},\quad\mbox{and}
	\\
	&\mathcal{N}_t:=[N]\setminus(\mathcal{S}_t\cup\mathcal{R}_t)=\{i:L_i^\mu(t)\leq \bar{\mu} \leq U_i^\mu(t)\}.
\end{align*}
Here, $\mathcal{R}_t$ and $\mathcal{S}_t$ can be thought of as empirical versions of $\mathcal{R}$ and $\mathcal{S}$ respectively. Next, we define the following events:
\begin{align*}
	&E_{ij}^\mu(t):=\{|\hat{\mu}_{ij}(t)-\mu_{ij}|\le\alpha(t, T_{ij}(t))\}, \label{Ei}\\
       &E_i^\mu(t):=\{|\hat{\mu}_i(t)-\mu_i|\le\alpha(t, T_{i}(t))\},\\
	&E_i(t):=E_i^\mu(t)\bigcap\bigg(\underset{j \in [M]}{\bigcap} E_{ij}^{\mathrm{\mu}}(t)\bigg),\quad \forall\, i\in[N] . 
\end{align*}
Further, for $t\geq 2$, we define
\begin{equation*}\label{E}
	E(t):=\bigcap_{i\in[N]} E_i(t)  \quad \mbox{and}\quad  E:=\bigcap_{t\ge 2} E(t) .
\end{equation*}
Note that as we pull of all attributes of an arm in any round of CSS-LUCB, we can conclude that $T_{ij}(t) = T_{i}(t)$. We now show that given our definition of the confidence radius $\alpha(t, T_i(t))$, the event $E$ occurs with high probability. First of all, using Hoeffding's inequality, we can claim that:
\begin{align*}
    \mathbb{P}[E_{ij}^\mu(t)^c] \leq \sum_{T_{ij}=1}^{t-1}2\cdot \exp(-2T_{ij}(t)\cdot\alpha(t,T_{ij})^2) = \frac{\delta}{2NMt^3} \\
       \mathbb{P}[E_{i}^\mu(t)^c] \leq \sum_{T_i=1}^{t-1}2\cdot \exp(-2T_{i}(t)\cdot\alpha(t,T_i)^2) = \frac{\delta}{2NMt^3}
\end{align*}
Using DeMorgan's laws and union bound, we can write:
\begin{align*}
   \mathbb{P}[E^c] \leq \mathbb{P}\Bigg[\underset{t\geq2}{\bigcup}\Bigg(\underset{i \in [N]}{\bigcup}\Bigg(E_i^\mu(t)^c\bigcup\bigg(\underset{j \in [M]}{\bigcup} E_{ij}^{\mathrm{\mu}}(t)^c\bigg)\Bigg)\Bigg)\Bigg] & \leq \sum_{t\geq2}\Bigg(\sum_{i \in [N], j\in[M]} \mathbb{P}[[E_{ij}^\mu(t)^c] + \sum_{i\in[N]} \mathbb{P}[E_{i}^\mu(t)^c]\Bigg) \\ & \leq \sum_{t\geq2}\Bigg(\sum_{i \in [N], j\in[M]} \bigg(\frac{\delta}{2NMt^3}\bigg) + \sum_{i \in [N]} \bigg(\frac{\delta}{2NMt^3}\bigg)\Bigg) \\ & \leq \sum_{t\geq2}\frac{\delta}{2t^3}\bigg(1 + \frac{1}{M}\bigg) \\ & \leq \frac{\delta}{8}\bigg(1 + \frac{1}{M}\bigg) \\ & \leq \frac{\delta}{4}.
\end{align*}
Thus, we conclude that event $E$ occurs with probability atleast $1 - \frac{\delta}{4}$. \\

\noindent Given the definitions of our empricial sets and the CSS-LUCB algorithm, we can proceed in a manner similar to Lemma 3 in \cite{va-lucb} to show that on the event $E(t)$, if CSS-LUCB does not terminate, then at least one of the following must hold: 

\begin{align}
    \label{condition1} i_{t} \in  (\partial\mathcal{F}_t\backslash\mathcal{S}_t)\cup(\mathcal{F}_t\cap \mathcal{N}_t)\\
    \label{condition2} c_{t} \in  (\partial\mathcal{F}_t\backslash\mathcal{S}_t)\cup(\mathcal{F}_t\cap \mathcal{N}_t).
\end{align}
This gives us a sufficient condition for the termination of the algorithm i.e. when neither of the arms $i_t$ or $c_t$ belong to the set $(\partial\mathcal{F}_t\backslash\mathcal{S}_t)\cup(\mathcal{F}_t\cap \mathcal{N}_t).$, the algorithm must have terminated. Our next aim is to show that after sufficiently many pulls of each arm (all its attributes), this set remains non empty with small probability. To begin with we define two quantities, for a sufficient large $ t $, let $ u_i(t) $ be the smallest number of pulls of a suboptimal arm $ i $ such that
$ \alpha(t,u_i(t)) $ is no greater than $ \Delta_i $, i.e.,
\begin{equation*}\label{ui}
	u_i(t):=\left\lceil{ \frac{1}{2 (\Delta_i)^2} \ln \left(\frac{4MN t^{4}}{\delta}\right)}\right\rceil
\end{equation*}
and $ v_i(t) $ be the smallest number of pulls of an arm $ i $ such that
$ \alpha(t,u_i(t)) $ is no greater than $ \Delta_i^{attr}$, i.e.,
\begin{equation*}\label{vi}
	v_i(t):=\left\lceil{\frac{1}{2 (\Delta_i^{attr})^2} \ln \left(\frac{4M N t^{4}}{\delta}\right)}\right\rceil
\end{equation*}

\begin{lem}\label{integral}
	The function $ h:\mathbb{R}\rightarrow\mathbb{R} $ defined by $$ h(x)=\exp\left(-2(\Delta_i^{\mathrm{v}})^2 \big(\sqrt{x}-\sqrt{v_i(t)} \big)^2\right) $$ is convex and decreasing on $ (4v_i(t),\infty) $ for all $ i\in[N] $. Furthermore, $$ \int_{4v_i(t)}^\infty h(x) \,\mathrm{d}x\leq \frac{\delta}{2(\Delta_{i}^{\mathrm{v}})^2 Nt^4}.$$
\end{lem}

\noindent Next, we find the following probabilities : 
\begin{itemize}
\item Probability that a suboptimal arm $i$ does not belong to the set $\mathcal{S}_t$, after being pulled more than $16u_i(t)$ times \item  Probability that the best arm $i^*$ does not belong to the set $\mathcal{R}_t$, after being pulled more than $16u_{i^*}(t)$ times \item Probability that a feasible arm $i$ does not belong to the set $\bar{\mathcal{F}_t}$, after being pulled more than $4v_i(t)$ times \item Probability that an infeasible arm $i$ does not belong to the set $\bar{\mathcal{F}_t}^c$, after being pulled more than $4v_i(t)$ times\\
\end{itemize}
\noindent (i)
\begin{align}
    \mathbb{P}[T_i(t)>16u_i(t), i\notin \mathcal{S}_t]
    &=\mathbb{P}[T_i(t)>16u_i(t), U_i^\mu(t)\geq\bar{\mu}]\notag \\
    &\leq\sum_{T=16u_i(t)+1}^\infty \mathbb{P}[T_i(t)=T,\hat{\mu}_i(t)-\mu_i\geq\bar{\mu}-\alpha(t,T)-\mu_i]\notag \\
    &\overset{(a)}\leq\sum_{T=16u_i(t)+1}^\infty \exp\left(-2T\left(\bar{\mu}-\alpha(t,T)-\mu_i\right)^2\right)\notag \\
    &\overset{(b)}\leq \sum_{T=16u_i(t)+1}^\infty \exp\left(-2T\left(\frac{\Delta_i}{2}-\sqrt{\frac{1}{2 T} \ln \left(\frac{4MN t^{4}}{\delta}\right)}\right)^2\right)\notag \\
    &\leq\sum_{T=16u_i(t)+1}^\infty \exp\left(-2\Delta_{i}^2\left(\frac{\sqrt{T}}{2}-\sqrt{u_i(t)}\right)^2\right)\notag \\
    &\overset{(c)}\leq\int_{16u_i(t)}^\infty \exp\left(-2\Delta_{i}^2\left(\frac{\sqrt{x}}{2}-\sqrt{u_i(t)}\right)^2\right)\,\mathrm{d}x\notag \\
    &=\int_{4u_i(t)}^\infty 4\exp\left(-2\Delta_{i}^2\left(\sqrt{x}-\sqrt{u_i(t)}\right)^2\right)\,\mathrm{d}x\notag \\
    & \overset{(d)}\leq \frac{4\delta}{4(\Delta_i)^2MNt^4}\notag \\
    &\leq\frac{\delta}{4(\frac{\Delta_{i}}{2})^2 MNt^4} \notag.
\end{align}
Here, (a) follows from Hoeffding's inequality. The quantity $\bar{\mu}-\alpha(t,T)-\mu_i$ is positive because for $T > 16u_i(t), \quad \alpha(t,T_i(t))$ can be at most $\Delta_i/4$, and $\mu_i$ and $\mu_i$ satisfy the following condition $\frac{\Delta_i}{2}\leq\bar{\mu}-\mu_i \leq \Delta_i$ for any suboptimal arm. Inequality (b) follows from the same condition. As the integrand is convex and decreasing the summation can be upper bounded by the integral as done in (c), and (d) follows from \ref{integral},\\

\noindent (ii) Similarly for the best arm $i^*$:
\begin{align*}
    \mathbb{P}[T_{i^*}(t)>16u_{i^*}(t), i^*\notin \mathcal{R}_t]
    &=\mathbb{P}[T_{i^*}(t)>16u_{i^*}(t), L_{i^*}^\mu(t)\leq\bar{\mu}]\\ &  \leq\sum_{T=16u_{i^*}(t)+1}^\infty \mathbb{P}[T_i(t)=T,\hat{\mu}_{i^*}(t)-\mu_{i^*}\leq -(\mu_{i^*} - \bar{\mu} - \alpha(t,T))] \\ & \overset{(a)}\leq\sum_{T=16u_{i^*}(t)+1}^\infty \exp\left(-2T\left(\frac{\Delta_{i^*}}{2}-\alpha(t,T)\right)^2\right) \\     &\overset{(b)}{\leq}\frac{\delta}{4(\frac{\Delta_{i^*}}{2})^2 MNt^4}.
\end{align*}
Here (a) uses the fact $\frac{\Delta_{i^*}}{2} = \mu_{i^*} - \bar{\mu}$ and (b) follows from arguments similar to case (i). Also note that we can claim the following:

\begin{align}
    &\label{final_inequality_1}\mathbb{P}[T_i(t)>16u_i(t), i\in \mathcal{N}_t] \leq \mathbb{P}[T_i(t)>16u_i(t), i\notin \mathcal{S}_t] \leq \frac{\delta}{4(\frac{\Delta_{i}}{2})^2 MNt^4} \\       &\label{final_inequality_2}\mathbb{P}[T_{i^*}(t)>16u_{i^*}(t), i^*\in \mathcal{N}_t] \leq   \mathbb{P}[T_{i^*}(t)>16u_{i^*}(t), i^*\notin \mathcal{R}_t] \leq \frac{\delta}{4(\frac{\Delta_{i^*}}{2})^2 MNt^4}.
\end{align}

\noindent (iii) Let $j' = \underset{j}{\text{argmin}} \ L_{ij} \ \forall i \in [N]$. Also, as the confidence radii for all the attributes of an arm are the same (in CSS-LUCB), we can claim that if $j'' = \underset{j}{\text{argmin}} \ \hat\mu_{ij}, \ j' = j''$. Define $j_{min} = \underset{j}{\text{min}} \ \mu_{ij}$. Now, for a feasible arm $i \in \mathcal{F}$, we can write:
\begin{align}
		\mathbb{P}[T_i(t)>4v_i(t),i\in \partial\mathcal{F}_t]
		&\leq\mathbb{P}[T_i(t)>4v_i(t),i\notin\mathcal{F}_t] \notag \\
		&=\mathbb{P}[T_i(t)>4v_i(t),L_{ij'} < \mu_{\text{TH}}] \notag \\
		&=\sum_{T=4v_i(t)+1}^\infty \mathbb{P}[T_i(t)=T, \hat{\mu_{ij'}} - \alpha(t, T) < \mu_{\text{TH}}]\notag \\
  &=\sum_{T=4v_i(t)+1}^\infty \mathbb{P}[T_i(t)=T, \hat{\mu_{ij'}} - \mu_{ij'} < - (\mu_{ij'} -\mu_{\text{TH}} - \alpha(t,T))] \notag \\   &\overset{(a)}\leq\sum_{T=4v_i(t)+1}^\infty \exp\left(-2T\left({\mu_{ij'}} - \mu_{\text{TH}} -\alpha(t,T)\right)^2\right)\notag \\ &\overset{(b)}\leq\sum_{T=4v_i(t)+1}^\infty \exp\left(-2T\left({\mu_{ij_{min}}} - \mu_{\text{TH}} -\alpha(t,T)\right)^2\right)\notag   \\ &\overset{(c)}\leq\sum_{T=4v_i(t)+1}^\infty \exp\left(-2T\left(\Delta_i^{attr} -\sqrt{\frac{1}{2 T} \ln \left(\frac{4MN t^{4}}{\delta}\right)}\right)^2\right)\notag  \\
    &\leq\sum_{T=4v_i(t)+1}^\infty \exp\left(-2(\Delta_{i}^{attr})^2\left(\sqrt{T}-\sqrt{v_i(t)}\right)^2\right)\notag \\&\leq\int_{4v_i(t)}^\infty \exp\left(-2(\Delta_{i}^{attr})^2\left(\sqrt{x}-\sqrt{u_i(t)}\right)^2\right)\,\mathrm{d}x\notag  \\
    &\label{final_inequality_3}\leq\frac{\delta}{4(\Delta_{i}^{attr})^2MNt^4}.
  \end{align}
  To arrive at (a), we again use Hoeffding's inequality. The term $\mu_{ij'} -\mu_{\text{TH}} - \alpha(t,T)$ is positive as $\mu_{ij'} -\mu_{\text{TH}} \geq \Delta_i^{attr}$, and for $T > 4v_i(t), \ \alpha(t,T)$ is at most $\Delta_i^{attr}/2$. Inequality (b) follows from the fact that $\mu_{ij_{min}} = \underset{j}{min} \  \mu_{ij}$ and (c) uses the fact that $\Delta_i^{attr} = \mu_{ij_{min}} - \mu_{\text{TH}}$. Finally, in (d), we again use Lemma \ref{integral} to get the upper bound on the probability.\\

\noindent  (iv) Keeping the definitions of $j'$ and $j_{min}$ the same, for an arm $i \in \mathcal{F}^c$, we can show that:
  \begin{align}
      \mathbb{P}[T_i(t)>4v_i(t),i\in \partial\mathcal{F}_t]
		&\leq\mathbb{P}[T_i(t)>4v_i(t),i\notin \mathcal{\bar{F}}_t^c] \notag \\
		&=\mathbb{P}[T_i(t)>4v_i(t), U_{ij} > \mu_{\text{TH}}] \quad (\forall j \in [M]) \notag\\
        & \overset{(a)} \leq
 \mathbb{P}[T_i(t)>4v_i(t), U_{ij_{min}} > \mu_{\text{TH}}] \notag\\ & \leq\sum_{T=4v_i(t)+1}^\infty \mathbb{P}[T_i(t)=T, \hat{\mu_{ij_{min}}} - \mu_{ij_{min}} >  (\mu_{\text{TH}} -\mu_{ij_{min}} - \alpha(t,T))] \notag \\     &\label{final_inequality_4}\overset{(b)}\leq\frac{\delta}{4(\Delta_{i}^{attr})^2MNt^4}.
 \end{align}
 In this case, (a) follows as the event $\{T_i(t)>4v_i(t), U_{ij} > \mu_{\text{TH}} \quad (\forall j \in [M])\}$ is a subset of the event $\{T_i(t)>4v_i(t), U_{ij_{min}} > \mu_{\text{TH}}\}$ and (b) is obtained by using arguments similar to previous cases.\\

\noindent Next using \ref{final_inequality_1}, \ref{final_inequality_2}, \ref{final_inequality_3} and \ref{final_inequality_4}, we show that after sufficiently many pulls, any of the arms belongs to the set.$(\partial\mathcal{F}_t\backslash\mathcal{S}_t)\cup(\mathcal{F}_t\cap \mathcal{N}_t)$ with low probability.
 \begin{itemize}
     \item For any feasible and suboptimal arm $i \in \mathcal{F} \cap \mathcal{S}$, after $T_i(t) > 16u_i(t)$, we showed in \ref{final_inequality_1} that $\mathbb{P}[i \in \mathcal{N}_t]$ as well as $\mathbb{P}[i \notin \mathcal{S}_t]$, with probability $\frac{\delta}{4(\frac{\Delta_{i}}{2})^2 MNt^4} $, and thus:
     \begin{align*}
         \mathbb{P}[i \in (\partial\mathcal{F}_t\backslash\mathcal{S}_t)\cup(\mathcal{F}_t\cap \mathcal{N}_t)] &\leq   \mathbb{P}[i \in (\partial\mathcal{F}_t\backslash\mathcal{S}_t)] +  \mathbb{P}[i \in (\mathcal{F}_t\cap \mathcal{N}_t)] \\ & \leq  \mathbb{P}[i \notin \mathcal{S}_t] +  \mathbb{P}[i \in \mathcal{N}_t] \\ & \leq \frac{\delta}{2(\frac{\Delta_{i}}{2})^2 MNt^4}.
      \end{align*}
      \item For the best arm $i^*$, note that if we show that $i^* \in (\mathcal{F}_t \cap \mathcal{R}_t)^c$ with low probability or in other words, it belongs to $(\mathcal{F}_t \cap \mathcal{R}_t)$ with high probability, it implies that it belongs to the set $(\partial\mathcal{F}_t\backslash\mathcal{S}_t)\cup(\mathcal{F}_t\cap \mathcal{N}_t)$ with low probability. Thus, for $T_{i^*}(t) > \text{max}(16u_{i^*}(t) , 4v_{i^*}(t))$ using \ref{final_inequality_2} and \ref{final_inequality_3}
    \begin{align*}
         \mathbb{P}[i^* \in (\partial\mathcal{F}_t\backslash\mathcal{S}_t)\cup(\mathcal{F}_t\cap \mathcal{N}_t)] &\leq  \mathbb{P}[i^* \in (\mathcal{F}_t \cap \mathcal{R}_t)^c] \\ & = \mathbb{P}[i^* \in (\mathcal{F}_t^c \cup \mathcal{R}_t^c)] \\ & \leq  \mathbb{P}[i^* \in \mathcal{F}_t^c] + \mathbb{P}[i^* \in \mathcal{R}_t^c] \\ & = \mathbb{P}[i^* \notin \mathcal{F}_t] + \mathbb{P}[i^* \notin \mathcal{R}_t] \\ & \leq \frac{\delta}{4(\Delta_{i^*}^{attr})^2 MNt^4} + \frac{\delta} {4(\frac{\Delta_{i^*}}{2})^2 MNt^4}. 
    \end{align*}

      \item For an infeasible and risky arm $i \in \mathcal{F}^c \cap \mathcal{R}$, after $T_i(t) > 4v_i(t)$, using \ref{final_inequality_4}
           \begin{align*}
         \mathbb{P}[i \in (\partial\mathcal{F}_t\backslash\mathcal{S}_t)\cup(\mathcal{F}_t\cap \mathcal{N}_t)] &\leq   \mathbb{P}[i \in (\partial\mathcal{F}_t\backslash\mathcal{S}_t)] +  \mathbb{P}[i \in (\mathcal{F}_t\cap \mathcal{N}_t)] \\ & \leq  \mathbb{P}[i \in \partial\mathcal{F}_t] +  \mathbb{P}[i \in \mathcal{F}_t] \\ & \leq  \mathbb{P}[i \in \partial\mathcal{F}_t] +  \mathbb{P}[i \notin \bar{\mathcal{F}_t^c}]\\ & \leq \frac{\delta}{2(\Delta_i^{attr})^2 MNt^4}.
         \end{align*}
    \item For an infeasible and suboptimal arm $i \in \mathcal{F}^c \cap \mathcal{S}$, if $16u_i(t) \geq 4v_i(t)$,
           \begin{align*}
         \mathbb{P}[i \in (\partial\mathcal{F}_t\backslash\mathcal{S}_t)\cup(\mathcal{F}_t\cap \mathcal{N}_t)] \leq\frac{\delta}{2(\Delta_i^{attr})^2 MNt^4}
         \end{align*}
        and if $16u_i(t) < 4v_i(t)$,
                   \begin{align*}
         \mathbb{P}[i \in (\partial\mathcal{F}_t\backslash\mathcal{S}_t)\cup(\mathcal{F}_t\cap \mathcal{N}_t)] \leq \frac{\delta}{2(\frac{\Delta_{i}}{2})^2 MNt^4}.
         \end{align*}
    Ignoring the ceiling operator, we can write
    \begin{align*}
         \mathbb{P}[i \in (\partial\mathcal{F}_t\backslash\mathcal{S}_t)\cup(\mathcal{F}_t\cap \mathcal{N}_t)] \leq \frac{\delta}{2\cdot\text{max}\big((\Delta_{i}^{attr})^2, (\frac{\Delta_{i}}{2})^2\big) MNt^4}.
         \end{align*}
 \end{itemize}
Let $t$ be a sufficiently large integer and $ \mathcal{T}:=\{\lceil t/2\rceil,\ldots, t-1\} $.
 Recall that the conditions \ref{condition1} and \ref{condition2} hold if and only if the event $E(t)$ occurs. The algorithm does not terminate at any $s \in \mathcal{T}$ either if the event $E(s)$ does not occur or if  $E(s)$ occurs and any of the arms belongs to the set $(\partial\mathcal{F}_s\backslash\mathcal{S}_s)\cup(\mathcal{F}_s\cap \mathcal{N}_s)$. We define the event $F_s = $ \{algorithm does not terminate at time s\}. The algorithm fails to terminate at time t if the conditions are not met at some $s \in \mathcal{T}$ and thus, using union bound, the probability that the algorithm does not terminate at time $t$ can be upper bounded as follows : 
\begin{align*}
    \mathbb{P}[F_t] &\leq \sum_{s\in\mathcal{T}} \mathbb{P}[F_s] \\ & \leq \sum_{s\in\mathcal{T}} \bigg(\mathbb{P}[E(t)^c] + \sum_{i\in[N]}   \mathbb{P}[i \in (\partial\mathcal{F}_t\backslash\mathcal{S}_t)\cup(\mathcal{F}_t\cap \mathcal{N}_t)]\bigg) \\& \leq \sum_{s\in\mathcal{T}} \bigg(\frac{\delta}{2s^3}\Big(1 + \frac{1}{M}\Big) + \frac{\delta}{2MNs^4}\Big(
    \sum_{i\in\mathcal{F} \cap \mathcal{S}} \frac{1}{(\frac{\Delta_{i}}{2})^2} + \sum_{i\in\mathcal{F}^c \cap \mathcal{R}}\frac{1}{(\Delta_i^{attr})^2} + \\ &\qquad \qquad \sum_{i\in\mathcal{F}^c \cap \mathcal{S}} \frac{1}{\text{max}\big((\Delta_{i}^{attr})^2, (\frac{\Delta_{i}}{2})^2\big)} +  \frac{1}{2(\Delta_{i^*}^{attr})^2} + \frac{1} {2(\frac{\Delta_{i^*}}{2})^2} \Big)\bigg) \\ & \leq \frac{2\delta}{t^2}\Big(1 + \frac{1}{M}\Big)  + \sum_{s\in\mathcal{T}}\bigg(\frac{\delta}{2MNs^4}\Big(
    \sum_{i\in\mathcal{F} \cap \mathcal{S}} \frac{1}{(\frac{\Delta_{i}}{2})^2} + \sum_{i\in\mathcal{F}^c \cap \mathcal{R}}\frac{1}{(\Delta_i^{attr})^2} + \\ &\qquad \qquad \sum_{i\in\mathcal{F}^c \cap \mathcal{S}} \frac{1}{\text{max}\big((\Delta_{i}^{attr})^2, (\frac{\Delta_{i}}{2})^2\big)} +  \frac{1}{2\cdot\text{min}\big((\Delta_{i^*}^{attr})^2, (\frac{\Delta_{i^*}}{2})^2\big)} + \frac{1} {2\cdot\text{min}\big((\Delta_{i^*}^{attr})^2, (\frac{\Delta_{i^*}}{2})^2\big)} \Big)\bigg) \\ & \overset{(a)}\leq \frac{2\delta}{t^2}\Big(1 + \frac{1}{M}\Big) + \frac{\delta\cdot H_{\text{id}}}{2MN}\sum_{s\in\mathcal{T}} \frac{1}{s^4} \\ & 
    \overset{(b)}\leq \frac{4\delta}{t^2} +  \frac{2\delta\cdot H_{\text{id}}}{t^3}.
\end{align*}
Here (a) follows from the definition of hardness index and (b) uses the fact that $M\geq1$ and $N\geq 2$. From our analysis, the additional number of time steps after $\left\lceil\frac{t}{2}\right\rceil -1$ required for the algorithm to terminate ($T_{max}$) are given by:
\begin{align*}
    T_{max} = \max\{16u_{i^\star}(t),4v_{i^\star}(t)\}
		+\sum_{i\in\mathcal{F}\cap\mathcal{S}}16u_i(t)
		&\;+\sum_{i\in\bar{\mathcal{F}}^c\cap\mathcal{R}} 4v_i(t)
		+\sum_{i\in\bar{\mathcal{F}}^c\cap\mathcal{S}}\min\{16u_i(t),4v_i(t)\}.
\end{align*}
If we chose $t = CH_{\text{id}}\ \text{ln}(H_{\text{id}}/\delta)$, this can further be upper bounded as follows:
\begin{align*}
    T_{max} &=\max\left\{16\left\lceil{\frac{1}{2 \Delta_{i^\star}^2} \ln \left(\frac{4MN t^{4}}{\delta}\right)}\right\rceil,4\left\lceil{\frac{1}{2 (\Delta_{i^\star}^{attr})^2} \ln \left(\frac{4M N t^{4}}{\delta}\right)}\right\rceil\right\}+
		16\sum_{i\in\mathcal{F}\cap\mathcal{S}}\left\lceil{\frac{1}{2 \Delta_i^2} \ln \left(\frac{4M N t^{4}}{\delta}\right)}\right\rceil\\
		&\quad +4\sum_{i\in\bar{\mathcal{F}}^c\cap\mathcal{R}}\left\lceil{\frac{1}{2 (\Delta_i^{attr})^2} \ln \left(\frac{4M N t^{4}}{\delta}\right)}\right\rceil
		+\sum_{i\in\bar{\mathcal{F}}^c\cap\mathcal{S}}\min\left\{16\left\lceil{\frac{1}{2 \Delta_{i}^2} \ln \left(\frac{4M N t^{4}}{\delta}\right)}\right\rceil,4\left\lceil{\frac{1}{2 (\Delta_{i}^{attr})^2} \ln \left(\frac{4M N t^{4}}{\delta}\right)}\right\rceil\right\} \\
		&\leq 16N+\max\left\{16{\frac{1}{2 \Delta_{i^\star}^2} \ln \left(\frac{4M N t^{4}}{\delta}\right)},4{\frac{1}{2 (\Delta_{i^\star}^{attr})^2} \ln \left(\frac{4MN t^{4}}{\delta}\right)}\right\}+
		16\sum_{i\in\mathcal{F}\cap\mathcal{S}}{\frac{1}{2 \Delta_i^2} \ln \left(\frac{4M N t^{4}}{\delta}\right)}\\*
		&\quad +4\sum_{i\in\bar{\mathcal{F}}^c\cap\mathcal{R}}{\frac{1}{2 (\Delta_i^{attr})^2} \ln \left(\frac{4M N t^{4}}{\delta}\right)}
		+\sum_{i\in\bar{\mathcal{F}}^c\cap\mathcal{S}}\min\left\{16{\frac{1}{2 \Delta_{i}^2} \ln \left(\frac{4M N t^{4}}{\delta}\right)},4{\frac{1}{2 (\Delta_{i}^{attr})^2} \ln \left(\frac{4MN t^{4}}{\delta}\right)}\right\} \\ & \leq 16N + 2H_{\text{id}}\ln \left(\frac{4MN t^{4}}{\delta}\right) \\ &\leq 16N + 2H_{\text{id}}\ \text{ln}(4) + 2H_{\text{id}}\ \text{ln}\left(\frac{N}{\delta}\right) + 2H_{\text{id}}\ \text{ln}(M) + 8H_{\text{id}}\ \text{ln}\left(CH_{\text{id}}\ \text{ln}\Big(\frac{H_{\text{id}}}{\delta}\Big)\right) \\ & \overset{(a)}\leq (16 + 2\ \text{ln}(4) + 8\ \text{ln}(C))H_{\text{id}} + 2H_{\text{id}}\ \text{ln}\left(\frac{N}{\delta}\right) + 2H_{\text{id}}\ \text{ln}(M) + 16 H_{\text{id}}\ \text{ln}\Big(\frac{H_{\text{id}}}{\delta}\Big) \\ & \overset{(b)}\leq (16 + 2\ \text{ln}(4) + 8\ \text{ln}(C) + 2 + 2 + 16)\ H_{\text{id}}\ \text{ln}\Big(\frac{H_{\text{id}}}{\delta}\Big) \\ &= (36 + 2\ \text{ln}(4) + 8\ \text{ln}(C))\ H_{\text{id}}\ \text{ln}\Big(\frac{H_{\text{id}}}{\delta}\Big) \\ & \overset{(c)}\leq \frac{1}{2}CH_{\text{id}}\ \text{ln}\Big(\frac{H_{\text{id}}}{\delta}\Big).
\end{align*}
To arrive at (a), we use the fact that $H_{\text{id}} < (H_{\text{id}}/\delta)$ and $(H_{\text{id}}/\delta) > \text{ln}(H_{\text{id}}/\delta)$. To obtain (b), we notice that for Bernoulli reward distributions $N < H_{\text{id}}$ as the expected rewards are between 0 and 1, and we also assume that $M \leq N$. Finally, we find a suitable value of $C$ which satisfies (d) and we find that the lowest integral value of $C$ for which $\frac{C}{2} \geq 36 + 2\ \text{ln}(4) + 8\ \text{ln}(C)$ is 159. Now, we observe that for $C\geq159$ i.e. $t\geq159H_{\text{id}}\ \text{ln}(H_{\text{id}}/\delta)$, we have $H_{\text{id}} < t$> Using this, we can further simplify the bound on the probability that the algorithm does not terminate:
\begin{align*}
    P[F_t] & \leq \frac{4\delta}{t^2} +  \frac{2\delta\cdot H_{\text{id}}}{t^3} \\ & \leq \frac{4\delta}{t^2} +  \frac{2\delta}{t^2} \\ &= \frac{6\delta}{t^2}.
\end{align*} 

\noindent  If the algorithm terminates, it succeeds on event $E$, which occurs with probability at least $1-{\delta}/{4}$. As shown above, the algorithm terminates at time 
    $t>159H_{\text{id}}\ \text{ln}(H_{\text{id}}/\delta)>6$ 
	with probability at least ${6\delta}/{t^2}$. So it succeeds after $O(H_\mathrm{id}\ln({H_\mathrm{id}}/{\delta}) )$ time steps with probability at least $1-({\delta}/{4}+{6\delta}/{6^2})\geq1-\delta$ and this completes the proof of upper bound.

\end{document}